\documentclass[12pt,letterpaper,reqno]{amsart}
\usepackage{amssymb,amsmath,amsfonts,amsthm,amsbsy,color}
\usepackage[utf8]{inputenc}
\usepackage{natbib}
\usepackage{graphicx}
\usepackage{epsfig,subfigure}
\usepackage{enumerate}
\usepackage{adjustbox}
\usepackage{units}
\usepackage{caption}
\usepackage{setspace,graphicx,setspace,multirow,lscape,longtable,dcolumn}
\usepackage{fullpage}
\usepackage[table, xcdraw]{xcolor}
\usepackage[capposition=top]{floatrow}
\usepackage{booktabs,caption}
\usepackage[flushleft]{threeparttable}
\usepackage{dcolumn}
\usepackage[normalem]{ulem}
\usepackage{arydshln}
\usepackage[algoruled,boxed,lined]{algorithm2e}
\usepackage{placeins}

\usepackage{color,soul}

\newcommand{\bs}{\boldsymbol}

\DeclareMathOperator*{\argmin}{arg\,min}

\newtheorem*{theorem*}{Theorem}

\graphicspath{ {figures/} }

\title{B\MakeLowercase{oo}ST: Boosting Smooth Transition Regression Trees for Partial Effect Estimation in Nonlinear Regressions}

\author{}
\begin{document}
\maketitle

\noindent\textbf{Yuri R. Fonseca}\newline
Graduate School of Business\newline
Columbia University \newline
E-mail: \texttt{yuri.fonseca@columbia.edu}
\vskip 1em
\noindent\textbf{Marcelo C. Medeiros}\newline
Department of Economics\newline
Pontifical Catholic University of Rio de Janeiro (PUC-Rio)\newline
E-mail: \texttt{mcm@econ.puc-rio.br}
\vskip 1em
\noindent\textbf{Gabriel F.R. Vasconcelos}\newline
Department of Economics\newline
University of California, Irvine \newline
E-mail: \texttt{gabriel.vasconcelos@uci.edu}
\vskip 1em
\noindent\textbf{Álvaro Veiga}\newline
Department of Electrical Engineering\newline
Pontifical Catholic University of Rio de Janeiro (PUC-Rio)\newline
E-mail: \texttt{alvf@ele.puc-rio.br}\newline
\newline
\noindent
\textbf{Abstract}:
In this paper, we introduce a new machine learning (ML) model for nonlinear regression called the \emph{Boosted Smooth Transition Regression Trees} (BooST), which is a combination of boosting algorithms with smooth transition regression trees. The main advantage of the BooST model is the estimation of the derivatives (partial effects) of very general nonlinear models. Therefore, the model can provide more interpretation about the mapping between the covariates and the dependent variable than other tree-based models, such as Random Forests. We present several examples with both simulated and real data.

\clearpage


\section{Introduction}

We introduce a new machine learning (ML) model for nonlinear regression called the \emph{Boosted Smooth Transition Regression Trees}, BooST. The idea of the model is to use a boosting algorithm to construct an additive nonparametric regression model consisting of a weighted sum of smooth transition regression trees (STR-Tree) as discussed in \citet{da2008tree}. Boosting is a greedy algorithm that iteratively combines weak learners (models) in order to form a stronger model. In the BooST framework, the weak learner is a STR-Tree model, which is a regression tree with soft splits instead of sharp (discrete) ones. Hence, the BooST model is a generalization of the Boosted Regression Trees \citep{buhlmann2002consistency}.

The main advantage of replacing hard splits with soft ones is that in the model can be differentiated with respect to the regressors and partial effects can be computed analytically, providing more interpretation on the mapping between the covariates and the dependent variable. The analysis of partial effects is very important in several fields, such as economics, engineering, operations research, and biology, among others. For example, a common problem in both economics and operations research is the estimation of elasticities and demand functions. See, for example, \citet{coglianese2017anticipation} or \citet{fisher2017competition}, who consider derivatives of sales with respect to prices, or \citet{schulte2015effects}, who consider the effects of a change in environmental temperatures on metabolism. The importance of going beyond average partial effects was recently pointed out by \citet{chernozhukov2018sorted}, who propose a percentile based method that provide a representation of heterogeneous effects.  

Partial effects can be equally recovered by other nonparametric models, such as neural networks (deep and shallow) or kernel regression. See \citet{liu2009estimating}, \citet{altonji2012estimating} or \citet{dai2016optimal} for recent discussions. The main advantage of the BooST approach is that the tree nature of the model makes it more scalable than traditional alternatives and less sensitive to dimensionality problems. The analytical computation of the derivatives is much more tractable in the BooST framework than in the new generation of deep networks. Furthermore, there are no theoretical results showing a consistency of partial effects estimators based on deep neural networks.

Although regression trees can consistently estimate general nonlinear mappings, they are also well-known for their instability, i.e., a small change in the data may have a big impact in the final model. Algorithms like boosting, as discussed in \citet{friedman2001greedy}, or Random Forests, as proposed by \citep{breiman2001random}, attenuates the instability problem by using a combination of trees. However, due to the sharpness of the splits in traditional trees, using these models to understand the relationship between variables is a difficult task \citep{ferreira2015analytics}. In this sense, recent developments make use of advanced algorithms in order to understand how changes to a specific decision variable affect the output of a tree-based model. See, for instance, \citet{imbens2017optimized} and \citet{mivsic2017optimization}.

\begin{figure}[htb]
	\centering
	\includegraphics[scale=0.5]{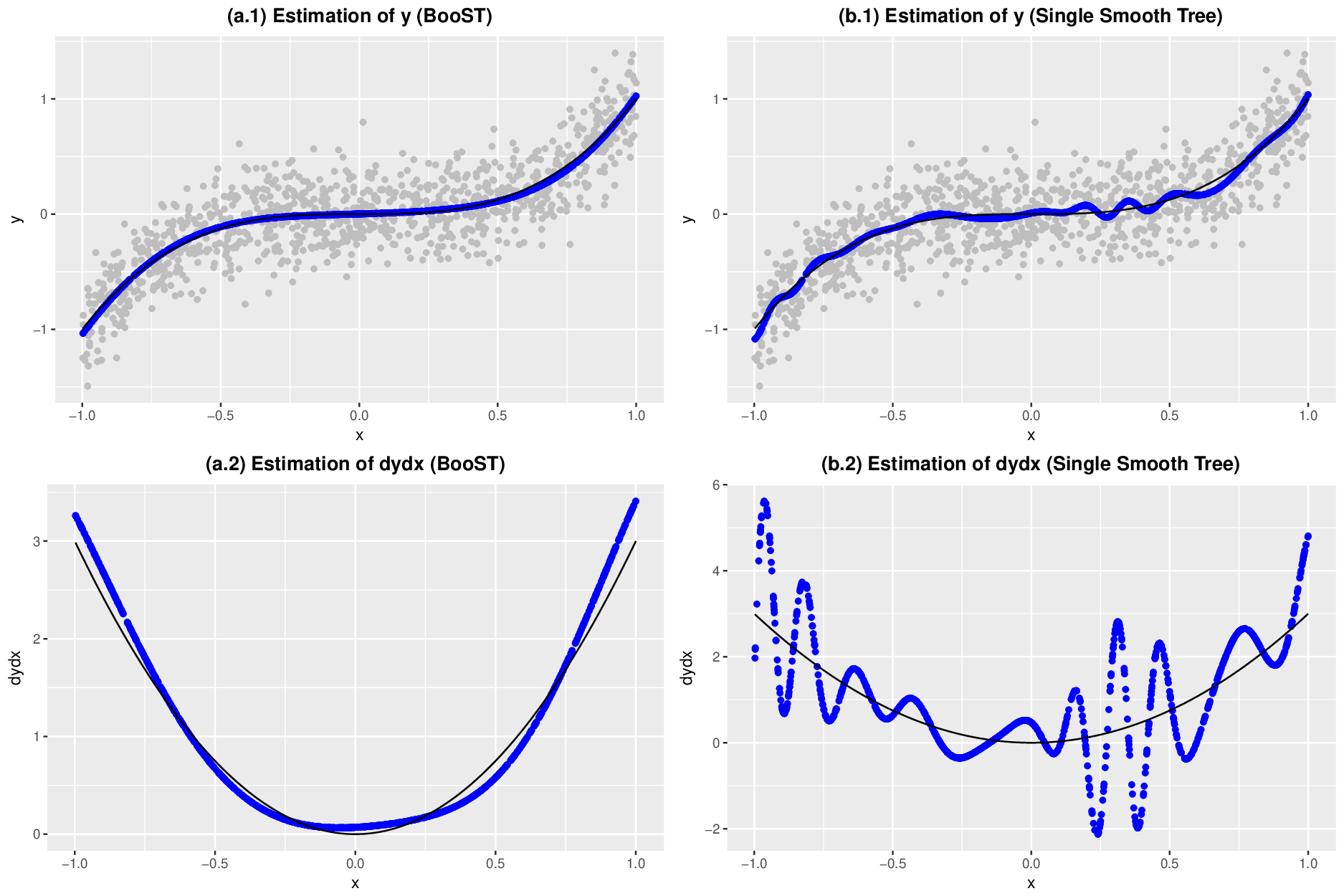}
	\caption{BooST and Smooth Tree example with the dgp $y_i=x_i^3+\varepsilon_i$ with $R^2$ set to 0.5. $x_i$ and $\varepsilon_i$ are two independently and normally distributed zero-mean random variables.}
	\label{F:basicexample}
\end{figure}

Random Forests and boosting differ in an important way: the first is usually estimated from large and independent regression trees with bootstrap samples, and the second is estimated iteratively with small trees in which each new tree is estimated on the pseudo residuals of the previous tree. We adopted the boosting algorithm because STR-Trees are computationally burdensome compared to sharp alternatives, which makes smaller trees more adequate than the large trees that are commonly used in Random Forests. Figure \ref{F:basicexample} shows how BooST performs compared to an individual STR-tree. The figure shows the fit of a single versus a boosted tree for the following data-generating process: $y_i=x_i^3+\varepsilon_i$, where $x_i$ and $\varepsilon_i$ are two independently and normally distributed zero-mean random variables with variance such that the $R^2$ of the mapping is set to to 0.5. Even for this simple example, one tree alone fails to produce reliable estimates of the derivatives.

This paper is organized as follows. We discuss trees and Smooth Transition Trees in Section 2. Section 3 presents the BooST algorithm. Examples using simulated data and empirical examples are given in Sections 4 and 5. Finally, our final remarks are provided in Section 6. The proofs are relegated to the Appendix.

\section{Regression Trees and Smooth Transition Trees}

\begin{figure}[htb]
	\centering
	\includegraphics[scale=0.5]{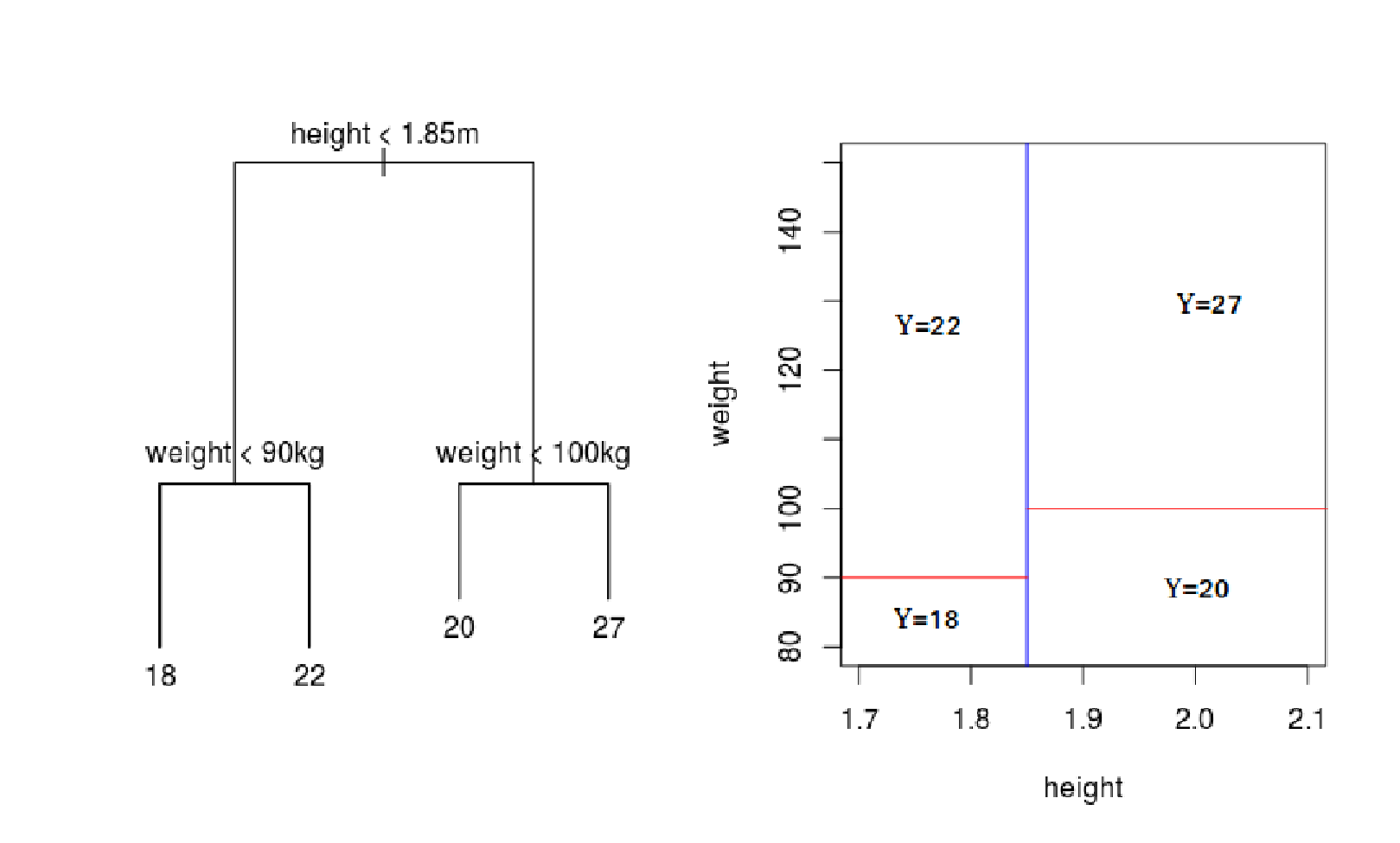}
	\caption{Example of a regression tree for basketball scores}
	\label{F:exampletree}
\end{figure}

A regression tree is a nonparametric model that approximates an unknown nonlinear function with local predictions using recursive partitioning of the space of the covariates. A tree may be represented by a graph, as in the left side of Figure \ref{F:exampletree}, which is equivalent to the partitioning in the right side of the figure in this bi-dimensional case. Suppose that we want to predict the scores of basketball players based on their height and weight. The first node of the tree in the example splits the players taller than 1.85 m from the shorter players. The second node on the left takes the short player groups and splits the players by weight, and the second node on the right does the same for the taller players. The prediction for each group is displayed in the terminal nodes, and they are calculated as the average score in each group. To grow a tree, we must find the optimal splitting point in each node, which consists of an optimal variable and an optimal cut-off. In the same example, the optimal variable in the first node is height, and the cut-off is 1.85 m.

STR-Trees differ from the usual trees as instead of assigning observations to nodes, they assign to observations the probability of belonging to a particular node. In the example depicted in Figure \ref{F:exampletree}, a regression tree states that a player taller than 1.85 m and heavier than 100 kg will be in the last node on the right. If this were a smooth tree, the same player would be more likely to be in the same node, but it would still be possible for him to be in other groups. Therefore, a STR-Tree can also be interpreted as a fuzzy regression tree. This simple change makes STR-Tree models differentiable with respect to the covariates, making it possible the estimation of partial effects. For example, in a STR-Tree model, one can estimate the variation in a player's score if he or she gained a little weight conditional on both his or her height and current weight. However, STR-Trees are more difficult to estimate and demand more computational power.

\subsection{Formal definition}

Let $\bs{x_i} = (x_{i,1},\dots,x_{i,m})'\in\mathbb{R}^m$ be an independently and identically distributed (IID) random vector of covariates and $y_i\in\mathbb{R}$ be a response (dependent) variable such that for $i=1,\ldots,N$:
\begin{equation}
\begin{split}
y_i&=\mathbb{E}(y_i|\bs{x_i}) + \varepsilon_i\\
y_i&=f(\bs{x_i}) + \varepsilon_i
\end{split}
\end{equation}
where $\{\varepsilon_i\}_{i=1}^N$ is an IID sequence of random variables with a zero mean.

A regression tree model with $K$ terminal nodes (leaves) approximates the function $f(\bs{x}_i)$ with an additive model $H(\bs{x}_i;\bs{\psi})$, which is indexed by the vector of parameters $\bs{\psi}$ and is constructed from a recursive partitioning of the space of covariates. $H(\cdot;\cdot)$ is a piece-wise constant function with $K$ subregions that are orthogonal to the axis of the predictor variables. Each subregion represents one terminal node. The tree also has $J$ parent nodes. For example, the tree in Figure \ref{F:exampletree} has three parent nodes ($J=3$) and four terminal nodes (regions) ($K=4$).

In order to represent any tree, we define the following notation. The root node is at position 0. Each parent node at position $j$ is split into two child nodes at position $2j+1$ and $2j+2$. Each parent node has a split (threshold) associated variable, $x_{s_j,i}\in\bs{x}_i$, where $s_j\in\mathbb{S}=\{1,2,\ldots,m\}$. Furthermore, $\mathbb{J}$ and $\mathbb{T}$ are the sets of parent and terminal nodes, respectively. For example, for the tree in Figure \ref{F:exampletree}, $\mathbb{J}=\{0,1,2\}$ and $\mathbb{T}=\{3,4,5,6\}$. The sets $\mathbb{J}$ and $\mathbb{T}$ uniquely identify the architecture (structure) of the tree.

Therefore,
\begin{equation}\label{E:model}
H(\bs{x}_i):=H_{\mathbb{J}\mathbb{T}}(\bs{x}_i;\bs{\psi})=\sum_{k\in\mathbb{T}}\beta_kB_{\mathbb{J}k}\left(\bs{x}_i;\bs{\theta}_k\right),
\end{equation}
where $0\leq B_{\mathbb{J}k}\left(\bs{x}_i;\bs{\theta}_k\right)\leq 1$ is a product of indicator functions defined as
\begin{equation}\label{E:B}
B_{\mathbb{J}k}\left(\bs{x}_i;\bs{\theta}_k\right)=
\prod_{j\in\mathbb{J}}I(x_{s_j,i};c_j)^{\frac{n_{kj}(1+n_{kj})}{2}}
\left[
1-I(x_{s_j,i};c_j)
\right]^{(1-n_{kj})(1+n_{kj})},
\end{equation}
with
\begin{equation}
I(x_{s_j,i};c_j)=
\begin{cases}
1 & \textnormal{if}\, x_{s_j,i}\leq c_j\\
0 & \textnormal{otherwise},
\end{cases}
\end{equation}
and
\begin{equation*}
n_{kj}=
\begin{cases}
-1 & \text{if the path to leaf } \,k\, \text{ does not include the parent node } j; \\
0  & \text{if the path to leaf } \,k\, \text{ include the right-hand child of the parent node } j; \\
1  & \text{if the path to leaf } \,k\, \text{ include the left-hand child of the parent node } j. \\
\end{cases}
\end{equation*}
where $c_j$ is the splitting point from a split in variable $j$. Note that the exponents in equation (\ref{E:B}) are either zero or one. The idea behind equation (\ref{E:B}) is to use the exponents to determine which path in the tree leads to each terminal node, if a node belongs (does not belong) to such path its exponent will be one (zero). Furthermore, define $\mathbb{J}_k$ as the set of indexes of parent nodes included in the path to leaf (region) $k$, such that $\bs{\theta}_k=\{c_j\}$ with $j\in\mathbb{J}_k$, $k\in\mathbb{T}$. Finally, it is clear that $\sum_{k\in\mathbb{T}}B_{\mathbb{J}k}\left(\bs{x}_i;\bs{\theta}_k\right)=1$.

\subsection{Introducing smoothness}

\begin{figure}[htb]
	\centering
	\includegraphics[scale=0.75]{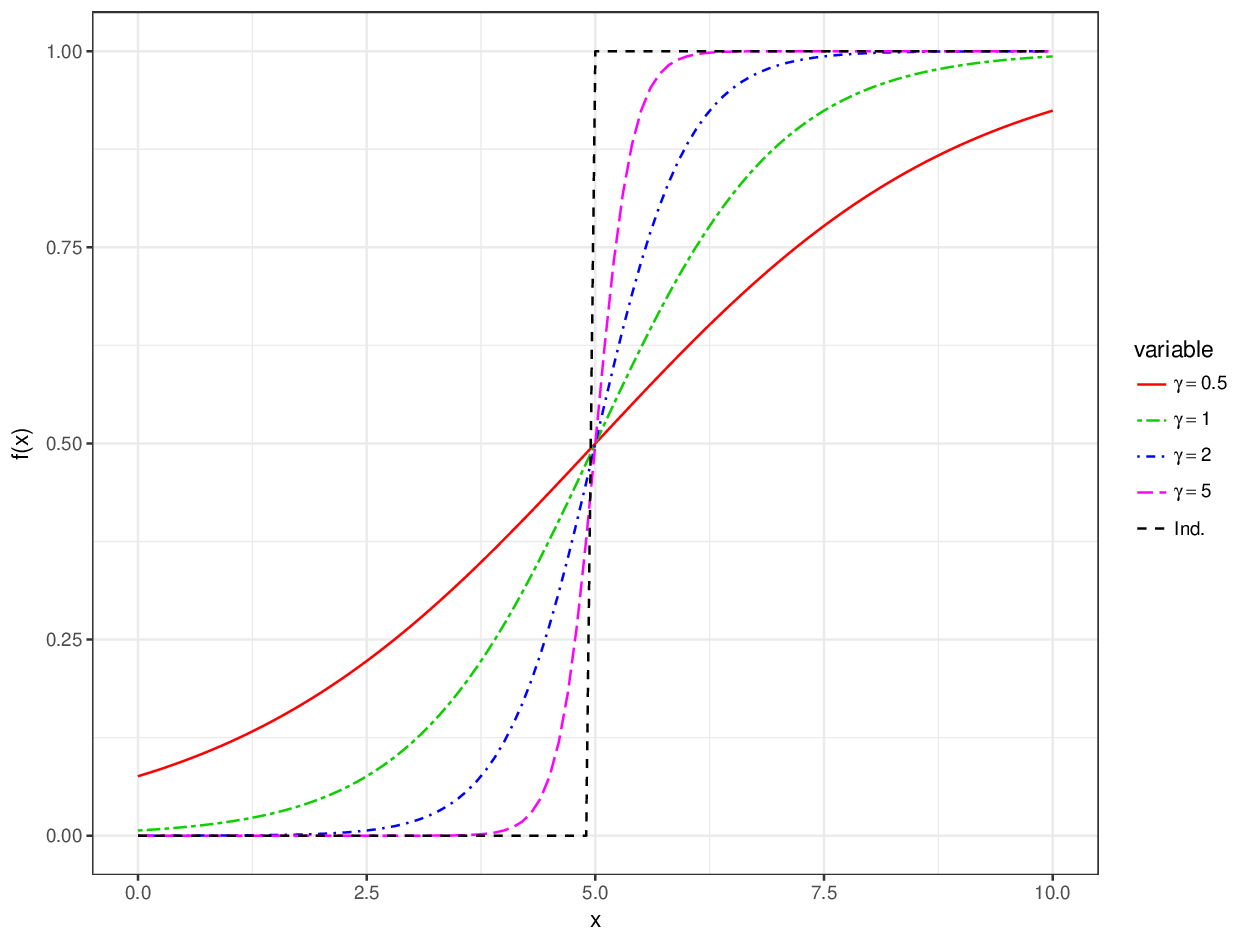}
	\caption{Logistic and indicator functions}
	\label{F:examplelogistic}
\end{figure}

In order to introduce smoothness, we follow \citet{da2008tree} and simply replace the discontinuous indicator function with a logistic function:
\begin{equation}
L(x_{s_j,i}; \gamma_j, c_j) = \frac{1}{1+e^{-\gamma_j(x_{s_j,i}-c_j)}},
\end{equation}
where $\gamma_j$ is the transition parameter, which controls the smoothness of the transition. The parameter $c_j$ is the location parameter. Figure \ref{F:examplelogistic} shows the indicator and logistic functions for several values of $\gamma$, setting $c = 5$. If $\gamma$ is very small, the logistic function becomes linear, and for large values of $\gamma$, the logistic becomes the indicator function.

The STR-Tree model is defined by equation (\ref{E:model}), with (\ref{E:B}) replaced by
\begin{equation}\label{E:Bmooth}
\begin{split}
B_{\mathbb{J}k}\left(\bs{x}_i;\bs{\theta}_k\right)&=
\prod_{j\in\mathbb{J}}B_{jk}(\bs{x}_i;\gamma_j,c_j)\\
&=\prod_{j\in\mathbb{J}}L(x_{s_j,i};\gamma_j,c_j)^{\frac{n_{k,j}(1+n_{kj})}{2}}
\left[
1-L(x_{s_j,i};\gamma_j,c_j)
\right]^{(1-n_{kj})(1+n_{kj})},
\end{split}
\end{equation}
where $\bs\theta_k=\{\gamma_j,c_j\}$, $j\in\mathbb{J}_k$.

In practical terms, the optimization above is broken in to smaller optimization problems. For each terminal node, variable and splitting point one must compute and store the squared error. Once this procedure is repeated for all nodes/variables/points, the solution will be the set of parameters that returns the smallest squared error. 

\subsection{Growing a Smooth Transition Regression Tree}

To grow a STR-Tree model, one must decide which node to split and the transition variable and estimate the smoothness and the location parameters. As in a regular tree, this process is done sequentially. Algorithm \ref{A:1} describes the overall procedure.

\begin{algorithm}
 \caption{Growing a Smooth Transition Regression Tree model}
 \label{A:1}
  \SetKwInOut{Input}{input}
  \SetKwInOut{Output}{output}
  \KwData{$\{x_i\}_{i=1}^N,\{y_i\}_{i=1}^N$}
  \Input{$\eta = 0$, $\mathbb{J}=\varnothing$, $\mathbb{T}=\varnothing$, and $K= \text{number of regions (terminal nodes)}$}
  \Output{$\{\beta_k\}_{k\in\mathbb{T}}, \{\bs{\theta}_k\}_{k\in\mathbb{T}}, \{\widehat{y}_i\}_{i=1}^N$}
  \While{$\eta < K$}{
  \eIf{$\eta = 0$}{
   Find the index of the best splitting variable, $s_0$, and splitting threshold $c_0$\;
   Compute $L(\bs{x}_{i,s_0};\gamma_0,c_0)$\;
   Compute $B_{\mathbb{J}k}(\bs{x}_i;\bs{\theta}_k)$, for $k = 1,2$\;
   Compute $\beta_k$, for $k = 1,2$\;
   Set $\mathbb{J}=\{0\}$ and $\mathbb{T}=\{1,2\}$\;
   }{
   Find the node to split, $j\in\mathbb{T}$, the index of the best splitting variable, $s_j$, the smoothness parameter, $\gamma_j$, and splitting threshold $c_j$ (location parameter)\;
   Compute $L(\bs{x}_{i,s_j};\gamma_j,c_j)$, $B_{\mathbb{J}k}(\bs{x}_i;\bs{\theta}_k)$, for $k = 2j+1,2j+2$\;
   Compute $\beta_k$, , for $k = 2j+1,2j+2$\;
   Update $\mathbb{J}$ and $\mathbb{T}$\;
  }
  Update $\eta=\eta+1$\;
 }
\end{algorithm}

The choice of the new node to split and the estimation of the parameters are done by minimizing the sum of the squared errors conditional on the previous splits. Consider a tree with terminal nodes $\mathbb{T}$ and suppose that we want to create a new split. We should simultaneously define which node $j\in\mathbb{T}$ should be split and the splitting variable and estimate both the smoothness and location parameters of the new split. Therefore,
\begin{equation}
\widehat{\bs\delta}\equiv(\widehat{j},\widehat{s}_j,\widehat{\gamma}_j,\widehat{c}_j,\widehat{\beta}_{2j+1},\widehat{\beta}_{2j+2})=\arg\min_{\bs\delta}\sum_{i=1}^N\left[y_i-Z(\bs{x}_i;\bs\delta|\mathbb{J},\mathbb{T})\right]^2,
\label{eq:opt}
\end{equation}
where
\begin{equation}
\begin{split}
Z(\bs{x}_i;\bs\delta|\mathbb{J},\mathbb{T})&=
\sum_{k\in\mathbb{T},k\neq j}\beta_kB_{\mathbb{J}k}\left(\bs{x}_i;\bs{\theta}_k\right)\\
&+\beta_{2j+1}L(x_{s_j,i};\gamma_j,c_j)B_{\mathbb{J}j}\left(\bs{x}_i;\bs{\theta}_j\right) + \beta_{2j+2}[1-L(x_{s_j,i};\gamma_j,c_j)]B_{\mathbb{J}j}\left(\bs{x}_i;\bs{\theta}_j\right).
\end{split}
\label{eq:optsplit}
\end{equation}

Although $j$ and $s_j$ are not parameters in the usual sense, they have to be estimated from the data. Therefore, we decided to include them in the definition of $\bs\delta$ in order to simplify notation. 

Growing a STR-Tree model is more complex than in the case of a traditional regression tree, where the search for the best splitting point becomes easier as the tree grows because we can look at terminal nodes individually, and the number of observations in each node decreases with the size of the tree. Therefore, we can grow separate branches of the tree in parallel. This feature does not hold for STR-Tree models. For every new node, we must look at all observations and all terminal nodes at the same time to find the best solution to the optimization problem. In other words, each decision affects the whole model. Thus, the STR-Tree model becomes more difficult to grow as we increase the number of terminal nodes to be tested.
\subsection{Derivatives}

The analytical derivatives of the STR-Tree model are straightforward to compute.

From equation (\ref{E:model}), it is easy to see that
\begin{equation}
\frac{\partial B_{\mathbb{J}k}(\bs{x}_i;\bs\theta_k)}{\partial x_{s,i}}=
\sum_{j\in\mathbb{J}}\left[\prod_{\ell\in\mathbb{J},\ell\neq j}B_{\ell k}(\bs{x}_i;\gamma_{\ell},c_{\ell})\right]\frac{\partial B_{jk}(\bs{x}_i;\gamma_{j},c_{j})}{\partial x_{s,i}},
\end{equation}
where
\begin{equation}
\frac{\partial B_{jk}(\bs{x}_i;\gamma_{j},c_{j})}{\partial x_{s,i}}=
\begin{cases}
0 & \textnormal{if $n_{kj}=-1$},\\
-\frac{\partial L(x_{s_j,i}; \gamma,c)}{\partial x_{s,i}}& \textnormal{if $n_{kj}=0$},\\
\frac{\partial L(x_{s_j,i}; \gamma,c)}{\partial x_{s,i}}& \textnormal{if $n_{kj}=1$},
\end{cases}
\end{equation}
and
\begin{equation}
    \frac{\partial L(x_{s_j,i}; \gamma,c)}{\partial x_{s,i}} =
\begin{cases}
\gamma L(x_{s,i}; \gamma,c) [1-L(x_{s,i}; \gamma,c)] & \textnormal{if $s_j=s$},\\
0 & \textnormal{if $s_j\neq s$}.
\end{cases}
\end{equation}

Finally,
\begin{equation}
\frac{\partial y_i}{\partial x_{s,i}}=\frac{\partial H_{\mathbb{J}\mathbb{T}}(\bs{x}_i;\bs\psi)}{\partial x_{s,i}}=\sum_{k\in\mathbb{T}}\beta_k\frac{\partial B_{\mathbb{J}k}(\bs{x}_i;\bs\theta_k)}{\partial x_{s,i}}
\end{equation}

\section{BooST}

\subsection{Motivation and a brief link with the literature}

The STR-Tree model suffers from the same instability issues as traditional (sharp) tree specifications. A small change in the data may result in very different trees, which makes the predictions and derivatives very unstable. However, the instability increases significantly if we look at the derivatives. For example, the results in Figure \ref{F:basicexample} are not very poor for an individual tree if we look at the fitted values for $y$. The derivatives, on the other hand, are completely unreliable.

One way to attenuate the instability problem is consider Random Forest models \citep{breiman2001random}. Random Forests use bootstrap samples to estimate fully grown trees and compute forecasts as the average forecast of all trees. \citet{breiman1996bagging} indicated that bootstrapping methods highly benefit from model instability to produce stable combined models. Random Forests also have randomness introduced in each tree by selecting the splitting variable among a randomly chosen subset of variables in each new split. However, since Smooth Trees are computationally more difficult to grow than regular trees, the Random Forest framework might impose some practical problems if one uses STR-Tree instead of CART models.

Boosting is another greedy method to approximate nonlinear functions that uses base learners for a sequential approximation. The model we use here, called Gradient Boosting, was introduced by \citet{friedman2001greedy} and can be seen as a Gradient Descendent method in functional space. Other boosting algorithms can also be considered Gradient Boosting, for example, \cite{mason2000boosting}.

The study of statistical properties of the Gradient Boosting is well developed. For example, for regression problems, \citet{duffy2002boosting} demonstrated bounds on the convergence of boosting algorithms using assumptions on the performance of the base learner. \citet{buhlmann2002consistency} shows results for consistency in the case of $\ell_2$ loss functions and three base models, which is the same framework we have in the BooST. \citet{buhlmann2003boosting} presented consistency and a minimax rate of convergence for the special case where the base leaner are smooth splines. These results, where generalized in \citet{bissantz2007convergence} for the case where the base learner are symmetric kernels. \cite{buehlmann2006boosting} proves consistency and convergence rates for in high dimensional models under sparsity assumptions and linear base learners. \citet{zhang2005boosting} proves convergence, consistency and results on the speed of convergence with mild assumptions on the base learners. Since boosting indefinitely leads to \textit{over-fitting} problems, some authors have demonstrated the consistency of boosting with different types of stopping rules, which are usually related to small step sizes, as suggested by \citet{friedman2001greedy}. Some of these works include boosting in classification problems and gradient boosting for both classification and regression problems. See, for instance, \citet{jiang2004process, lugosi2004bayes, bartlett2007adaboost, zhang2005boosting, yao2007early}.

\subsection{General approach}

Following \cite{zhang2005boosting}, let $\mathcal{F} = \{F:\mathbb{R}^p \rightarrow \mathbb{R}\}$ be a set of real value functions. We want to find a function $F^\star$ such that

\begin{equation}
  \mathcal{L}(F^\star) = \inf_{F \in \mathcal{S}} \mathcal{L}(F),
\end{equation}
where the functional $\mathcal{L}:F\rightarrow \mathbb{R}$ is convex.

For statistical problems such as regression or classification, in order to estimate $F^\star$, one can proceed as in \cite{friedman2001greedy}, defining the following form for $\mathcal{L}(F)$:

\begin{equation}
\mathcal{L}(F) = \mathbb{E}_{\bs{y},\bs{x}}\{L[\bs{y}, F(\bs{x})]\},
\label{total_risk}
\end{equation}
where $\mathbb{E}_{\bs{y},\bs{x}}$ is the expectation with respect to the joint distribution of $(\bs{y},\bs{x})$ and $L$ is a specific loss function that is convex on the second argument, such as the square loss function.

Therefore, we seek to solve the following optimization problem:

\begin{equation}
F^\star = \argmin_{F \in \mathcal{F}} \mathbb{E}_{\bs{y},\bs{x}}\{L[\bs{y}, F(\bs{x})]\} = \argmin_{F \in \mathcal{F}} \mathbb{E}_{\bs{y}}\{\mathbb{E}_{\bs{y}}\{L[\bs{y}, F(\bs{x})]|\bs{x}\}\},
\label{objective_boosting}
\end{equation}
where $F$ is restricted to an additive expansion of the form:
\begin{equation}
F(\bs{x})\equiv F(\bs{x};\{\rho_m,\bs\psi_m\}_{m=1}^M) = \sum_{m = 1}^M \rho_m h(\bs{x};\bs\psi_m).
\label{expansion}
\end{equation}
where $h$ is a base learner. In the present case,
\[
\begin{split}
h(\bs{x};\bs\psi_m)&:=H_{\mathbb{J}_m\mathbb{T}_m}(\bs{x};\bs\psi_m)\\
&=\sum_{k\in\mathbb{T}_m}\beta_{km}B_{\mathbb{J}_mk}(\bs{x};\bs\theta_{km}),
\end{split}
\]
where
\[
B_{\mathbb{J}_{km}}(\bs{x};\bs\theta{km})=\prod_{j\in\mathbb{J}_m}L(x_{s_j};\gamma_{jm},c_{jm})^{\frac{n_{kj}(1+n_{kj})}{2}}\left[1-L(x_{s_j};\gamma_{jm},c_{jm})\right]^{(1-n_{kj})(1+n_{kj})}.
\]

Using the empirical risk approximation for (\ref{total_risk}) we get
\begin{equation}
R_F(\bs{y},\bs{x}) = \frac{1}{N}\sum^N_{i = 1}L[y_i,F(\bs{x}_i)|\bs{x}_i].
\label{ER}
\end{equation}

In order to find $F$ that minimize (\ref{ER}), we will follow the greedy approach proposed by \cite{friedman2001greedy}. The Gradient Boosting algorithm is based in the \textit{steepest-descent} algorithm, where at each iteration of the algorithm, we take a step in the opposite direction of the gradient of the loss function $L$ evaluated in the finite samples.

The algorithm works as follows. Assuming that we have computed the algorithm until iteration $m-1$, the gradient at the $m$-th iteration is calculated as
\begin{equation*}
u_m(\bs{x}_i) = \frac{\partial \mathbb{E}_{\bs{y},\bs{x}}\{L[y_i,F(\bs{x}_i)]\}}{\partial F(\bs{x}_i)}\bigg{|}_{F(\bs{x}_i) = F_{m-1}(\bs{x}_i)}.
\end{equation*}
Assuming the appropriate regularity conditions, we can re-write
\begin{equation*}
u_m(\bs{x}_i) = \frac{\partial R_F(\bs{y},\bs{x}_i)}{\partial F(\bs{x}_i)}\bigg{|}_{F(\bs{x}_i) = F_{m-1}(\bs{x}_i)},
\end{equation*}
where
\begin{equation*}
F_{m-1}(\bs{x}_i) = \sum_{j=0}^{m-1}\rho_jh_j(\bs{x}_i).
\end{equation*}

Using a two-step procedure, we first solve
\begin{equation}
\bs\psi_m = \argmin_{{\bs\psi}} \sum_{i=1}^N \left[-u_m(\bs{x}_i) - h(\bs{x}_i;\bs\psi)\right]^2.
    \label{objective_boosting_3}
\end{equation}
Then, we compute the step size or line search in the direction of $-u_m$ by solving
\begin{equation}
\rho_m = \argmin_\rho\sum_{i=1}^N L[y_i,F_{m-1}(\bs{x}_i) + \rho h(\bs{x}_i;\bs\psi_m)].
\label{opt_rho}
\end{equation}
Finally, the updated model at the $m$-th step will be given by
\begin{equation}
    F_m(x) = F_{m-1}(x) + \rho_m h(\bs{x}_i,\bs\psi_m),
    \label{update}
\end{equation}
and the predictions of the final model will be given by
\begin{equation}
\widehat{y}_i = \widehat{F}_{M}(\bs{x}_i) = \widehat{F}_0 + \sum_{m=1}^{M} \rho_m h(\bs{x}_i,\widehat{\bs\psi}_m),
\label{predictions}
\end{equation}
where $M$ is the total number of base learners and $F_0$ is the initial estimation. Another factor that is commonly used in Gradient Boosting is the addition of a \textit{shrinkage} parameter $v$ in equation (\ref{update}). Hence, the updated equation and prediction will be given by
\begin{equation*}
F_m(\bs{x}_i) = F_{m-1}(\bs{x}_i) + v\rho_m h(\bs{x}_i,\bs\psi_m),
\end{equation*}
and
\begin{equation*}
\widehat{y}_i = \widehat{F}_{M}(\bs{x}_i) = \widehat{F}_0 + \sum_{m=1}^{M} v\rho_mh(\bs{x}_i,\widehat{\bs\psi}_m).
\end{equation*}

It is worth noting that the parameter $v$ is not used in the estimation of (\ref{objective_boosting_3}) and (\ref{opt_rho}). However, theoretical and empirical results show that this parameter is necessary for both convergence and consistency of the Gradient Boosting; see, for example, \citet{zhang2005boosting}.

\subsection{Algorithm}

Algorithm (\ref{A:2}) presents the simplified BooST model for a quadratic loss. It is recommended to use a shrinkage parameter $v\in (0,1]$ to control the learning rate of the algorithm. If $v$ is close to 1, we have a faster convergence rate and a better in-sample fit. However, we are more likely to have over-fitting and produce poor out-of-sample results. Additionally, the derivative is highly affected by over-fitting, even if we look at in-sample estimates. A learning rate between 0.1 and 0.2 is recommended to maintain a reasonable convergence ratio and to limit over-fitting problems.
\begin{algorithm}
 \caption{BooST}
 \label{A:2}
 initialization $\phi_{i0} = \bar{y}:=\frac{1}{N}\sum_{i=1}^Ny_i$\;
 \For{m=1,\dots,M}{
  make $u_{im} = y_i-\phi_{im-1}$\;
  grow a STR-Tree to fit $u_{im}$, $\widehat{u}_{im} =  \sum_{k\in\mathbb{T}_m}\widehat{\beta}_{km} B_{\mathbb{J}_mk}(\bs{x}_i;\widehat{\bs{\theta}}_{km})$\;
  make $\rho_m = \text{arg min}_\rho \sum_{i=1}^N[u_{im}-\rho\hat{u}_{im}]^2$\;
  update $\phi_{im} = \phi_{m-1i}+v\rho_m\hat{u}_{im}$\;
 }
\end{algorithm}

We made two adaptations to the STR-Tree model in order to improve the benefits of the boosting algorithm. First, the transition parameter $\gamma$ is randomized at a given interval for each new node in each tree. If variables have a different scale, we divide the randomized $\gamma$ by the variable standard deviation. The second modification concerns finding the best splitting variable in each new node. It is common in Random Forests and boosting to test only a fraction of the total number of variables randomly selected in each node. This approach is another type of shrinkage, which makes models more robust to over-fitting and makes estimation computationally faster because fewer variables are tested in each new node. We adopted the same strategy in the BooST algorithm. Using 50\% of the variables to grow each new node is sufficient to benefit from this strategy. In most of our examples, we used two-thirds of the candidate variables in each node.

The BooST fitted value may be written as
\begin{equation}
\begin{array}{cl}
    \widehat{y}_i &= \bar{y}+ \sum_{m=1}^M v\rho_m\widehat{u}_{im}\\
     & =  \bar{y}+ \sum_{m=1}^M v\widehat{\rho}_m \sum_{k\in\mathbb{T}_m}\widehat{\beta}_{km} B_{\mathbb{J}_mk}(\bs{x}_i;\widehat{\bs{\theta}}_{km})
\end{array}
\end{equation}
and the derivative with respect to $x_{s,i}$ will be
\begin{equation}
    \frac{\partial y_i}{\partial x_{s,i}}   = \sum_{m=1}^M v\widehat{\rho}_m \sum_{k\in \mathbb{T}_m}\widehat{\beta}_{km} \frac{\partial B_{\mathbb{J}_mk}(\bs{x}_i;\widehat{\bs{\theta}}_{km})}{\partial x_{s,i}}.
\end{equation}

\section{Examples with Simulated Data}

In this section, we consider several simulated data-generating process (DGP) to evaluate how BooST behaves for different data structures and parameter values. All DGPs were adapted from \citet{wager2014asymptotic}. We start with a DGP with only two variables, which provides visual examples of the BooST for different parameters. The data were generated with the following equation:
\begin{equation}
y_i = \cos[\pi(x_{1,i}+x_{2,i})]+\varepsilon_i,
\label{eq:dgp1}
\end{equation}
where $x_{1,i}\sim N(0,1)$, $x_{2,i}\sim Bernoulli(0.5)$ and $\varepsilon_i\sim N(0,\sigma^2)$, with $\sigma$ adjusted to obtain a pre-defined $R^2$ value.

\begin{figure}[htb]
	\centering
	\includegraphics[scale=0.7]{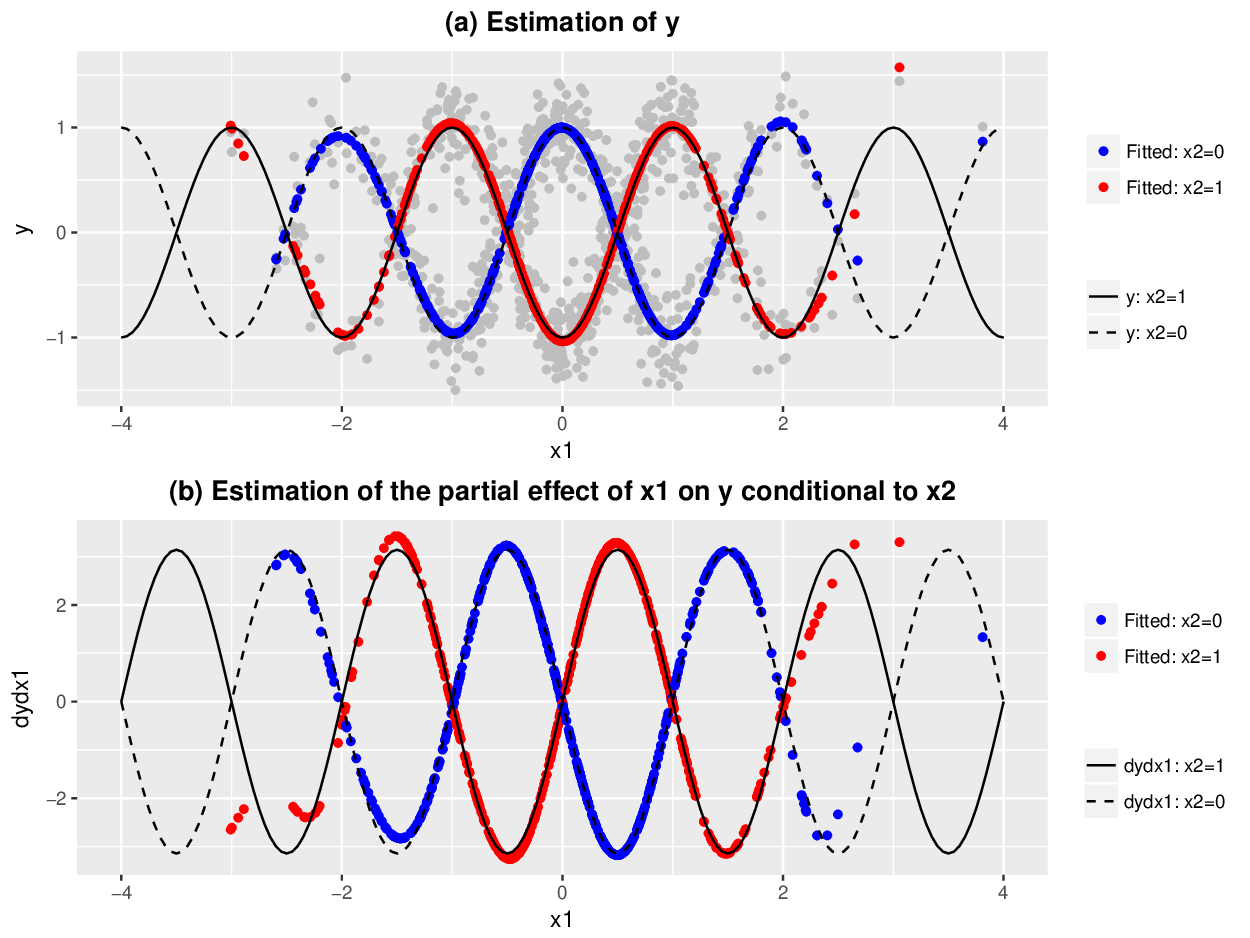}
	\caption{Example of a single estimation of the cosine DGP with $R^2=0.9$.}
	\label{F:excosine09}
\end{figure}

\begin{figure}[htb]
	\centering
	\includegraphics[scale=0.7]{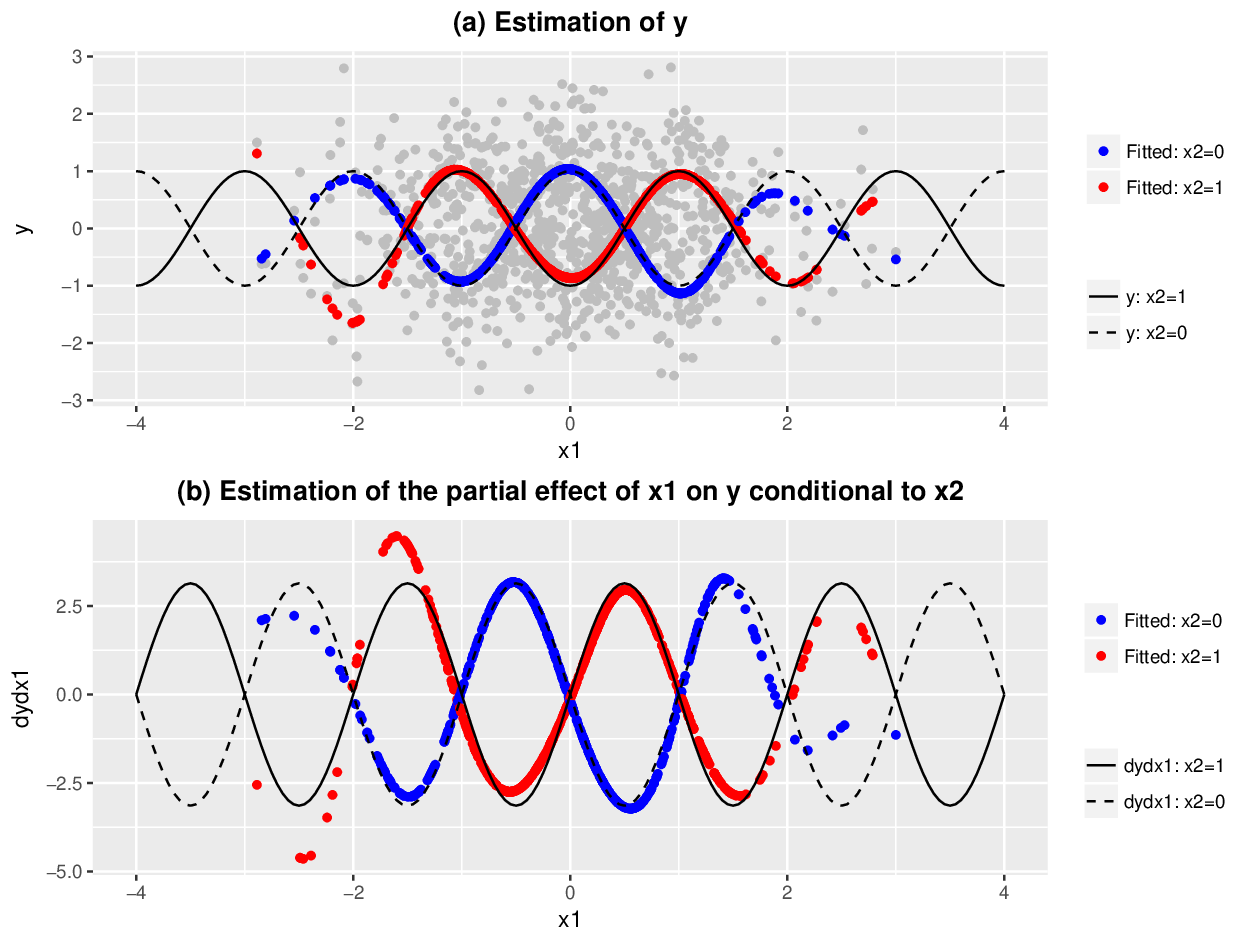}
	\caption{Example of a single estimation of the cosine DGP with $R^2=0.5$.}
	\label{F:excosine05}
\end{figure}

We will begin with two simple examples using data generated from (\ref{eq:dgp1}), with $R^2$ set to 0.9 and 0.5 and 1000 observations. We estimated the BooST model with $M=1000$ trees, $\gamma$ randomized in the $[0.5,5]$ interval, and each tree had four splits. Figure \ref{F:excosine09} shows the results for $R^2=0.9$. The fitted level is in panel (a), and the derivative is in panel (b). The generated data are displayed in gray dots in the plots. The model is very precise both for the fitted values and for the derivative. Figure \ref{F:excosine05} shows a more challenging design with the data showing much less structure. The results are still satisfactory, but we see some poor fit for the derivatives in the tails of the distribution of $x_1$. Since $x_1$ was generated from a normal distribution, it has fewer observations in the tails than in the middle quantiles, and the derivatives are more precise when the data are less sparse. This feature is normal for tree-based models. However, the derivative is more strongly affected by the lack of data in the tails than the levels. Still, the results are very good, given that the model is estimated with absolutely no knowledge of the nonlinear function that generated $y$.

In the next subsections, we will discuss convergence and parametrization for several model specifications. In each example, we will keep all parameters fixed, except by the one we wish to analyze. All examples were performed on data from the DGP in (\ref{eq:dgp1}) with $R^2=0.5$. The base model is the model in figure $\ref{F:excosine05}$.

\subsection{Convergence and shrinkage}

\begin{figure}[htb]
	\centering
	\includegraphics[scale=0.8]{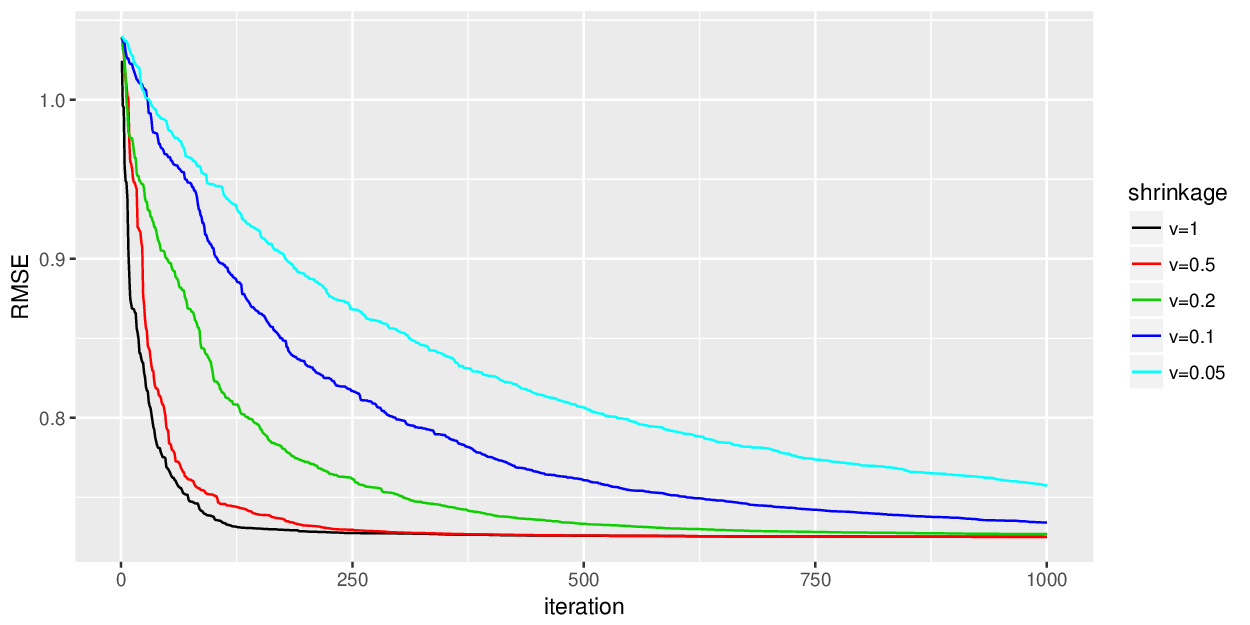}
	\caption{Convergence speed for different values of shrinkage ($v$).}
	\label{F:convergencecosinev}
\end{figure}

The first feature we analyzed was the trade-off between convergence and shrinkage. Figure \ref{F:convergencecosinev} shows the BooST convergence for several values of shrinkage $v$. The $y$ axis shows the root mean squared error of the adjusted model in each BooST iteration. Models with $v=0.5$ and $1$ converge with less than 250 iterations, the models with $v=0.2$ and $v=0.1$ take approximately 750 and 1000 iterations to converge, respectively, and the model with $v=0.05$ does not reach convergence with 1000 iterations.

\begin{figure}[htb]
	\centering
	\includegraphics[scale=0.5]{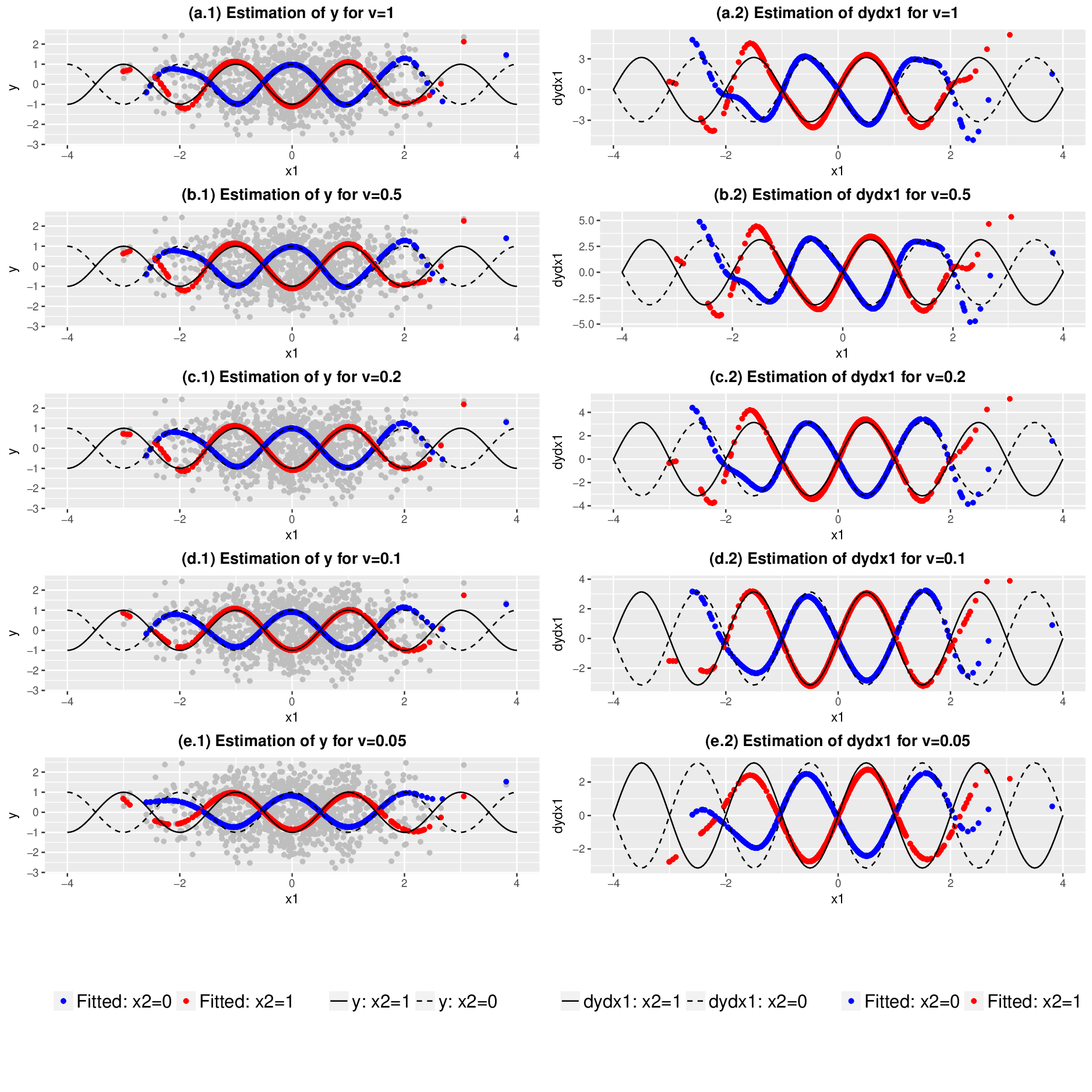}
	\caption{Fitted model for different values of shrinkage ($v$).}
	\label{F:fitsforvs}
\end{figure}

Naturally, large values of $v$ result in faster convergence. However, it is expected that the fitted model becomes noisier with larger values of $v$. The fitted values and the derivative for each $v$ is shown in figure \ref{F:fitsforvs}.

The left panels show the fitted values, and the right panels show the derivatives. Although the fitted values become slightly softer as we decrease $v$, the difference from $v=1$ and $v=0.05$ is small. However, if we look at the derivatives in the right column of the plots, it is possible to see the improvement as we decrease $v$ to 0.1. The model for $v=0.05$ did not converge, and the derivative curve did not completely reach the real values of the derivatives.

\subsection{Number of splits in each tree}

\begin{figure}[htb]
	\centering
	\includegraphics[scale=0.8]{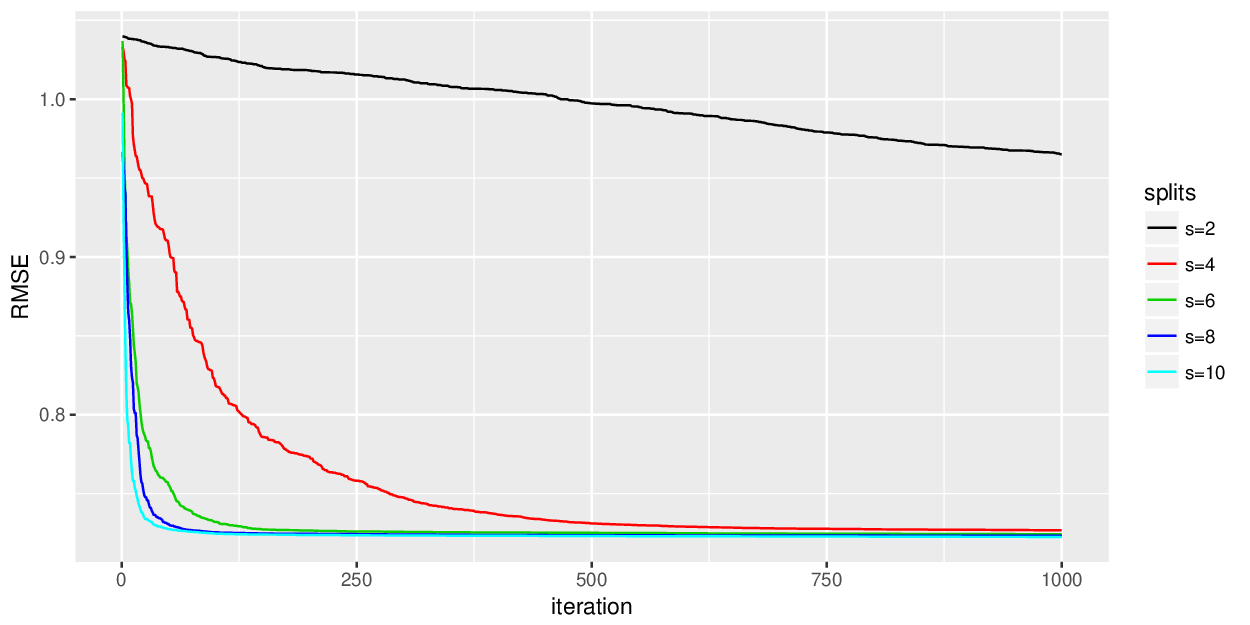}
	\caption{Convergence speed for different numbers of splits ($s$).}
	\label{F:convergencecosines}
\end{figure}

The next parameter to be analyzed is the number of splits in each tree. Figure \ref{F:convergencecosines} shows the convergence speed for several tree sizes. The convergence is faster as we increase the number of nodes in the trees, but the difference between 6, 8 and 10 splits is very small. However, the model with only two splits is very far from convergence after 1000 iterations of the BooST algorithm.

\begin{figure}[htb]
	\centering
	\includegraphics[scale=0.5]{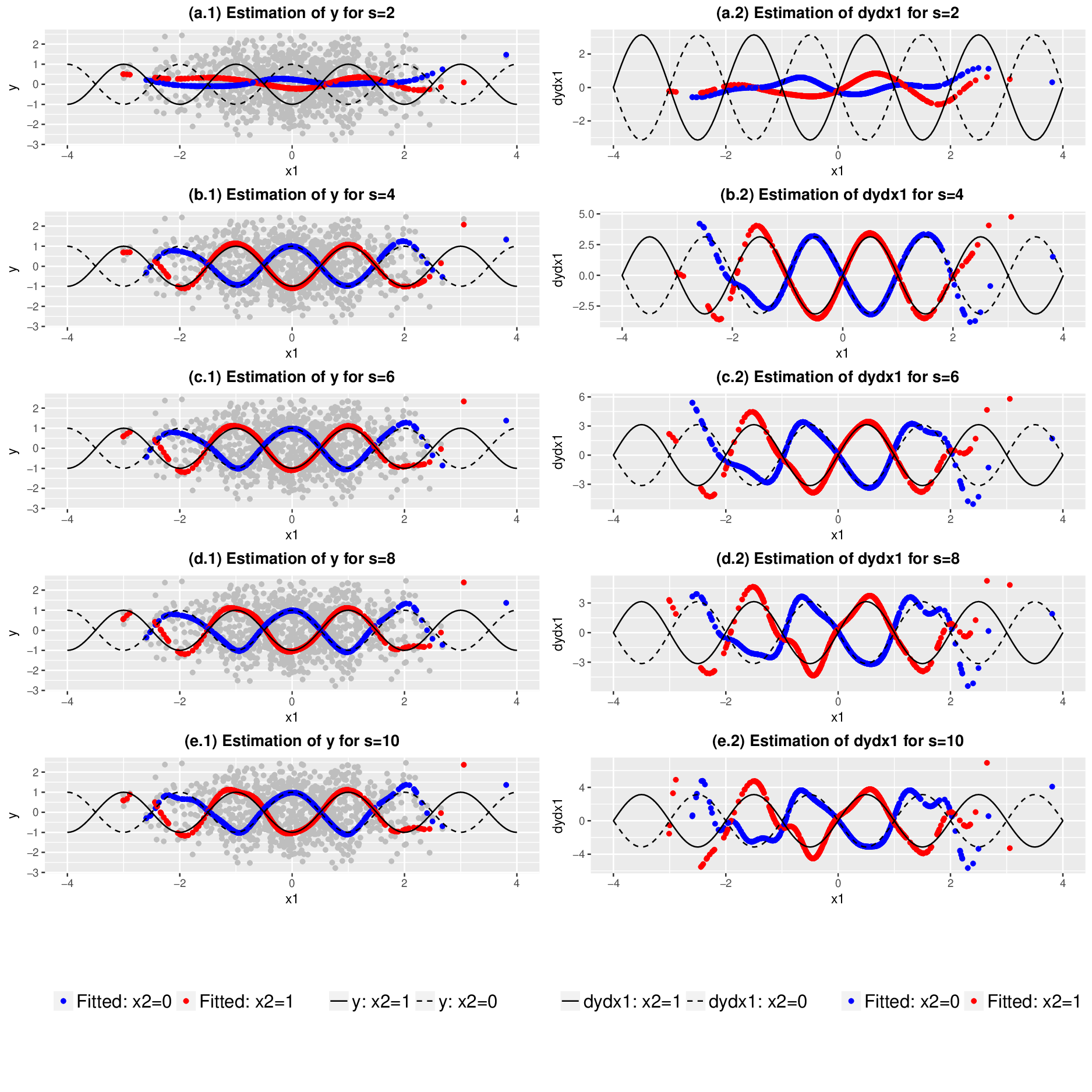}
	\caption{Fitted model for different numbers of splits ($s$).}
	\label{F:fitsforss}
\end{figure}

The fitted model and its derivatives are presented in figure \ref{F:fitsforss}. The fitted values on the left are very similar across models, except the model with two splits that did not reach convergence. However, we can see a significant deterioration of the derivative as we increase the number of splits in each tree for values greater than four.

\subsection{Range of the transition parameter $\gamma$}

\begin{figure}[htb]
	\centering
	\includegraphics[scale=0.8]{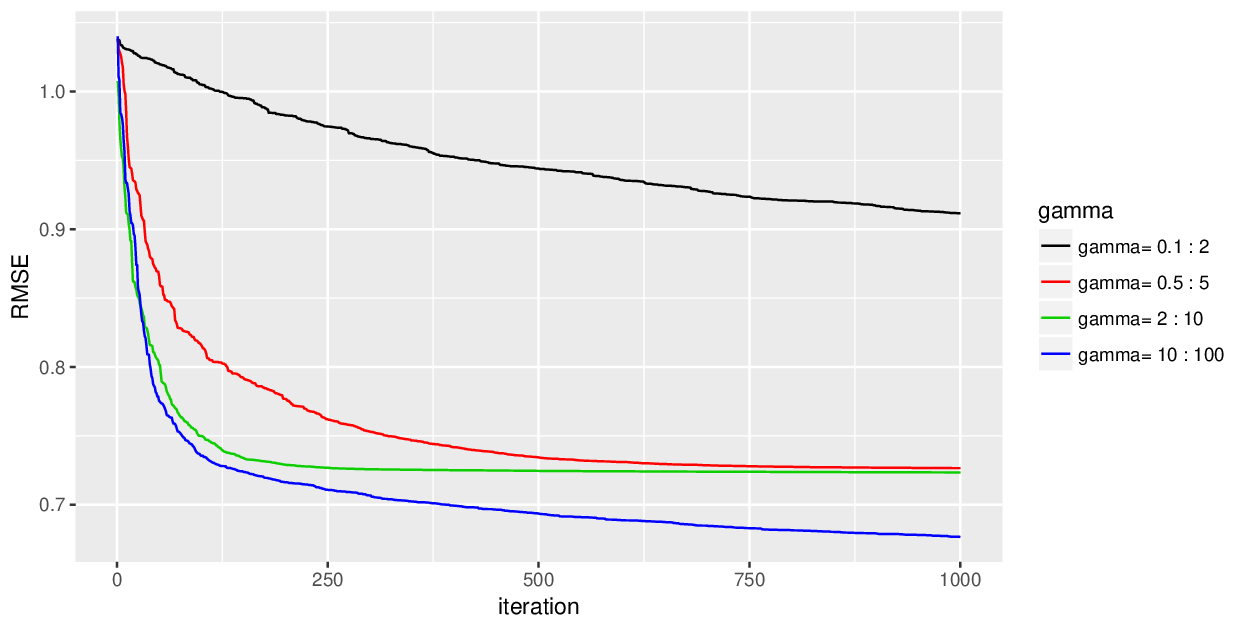}
	\caption{Convergence speed for different ranges of $\gamma$.}
	\label{F:convergencecosineg}
\end{figure}

The last parameter we analyzed is the range of $\gamma$s from which we sample the transition parameter in each node. This example is the most important one because it shows how BooST changes as we change the smoothness in the trees. The convergence for several ranges of $\gamma$s is shown in figure \ref{F:convergencecosineg}. Very small values of $\gamma$ introduce a great deal of smoothness in the tree, and convergence becomes more difficult. Recall from figure \ref{F:examplelogistic} that small values of $\gamma$ result in a close to linear relationship between the logistic function and $x$. Values of $\gamma$ in the $[0.5,5]$ and $[2,10]$ intervals produce similar results in terms of convergence. The interesting case is for $\gamma$ between 10 an 100. In this case, the model breaks the convergence barrier from all models presented so far, resulting in a much smaller in-sample RMSE.

\begin{figure}[htb]
	\centering
	\includegraphics[scale=0.5]{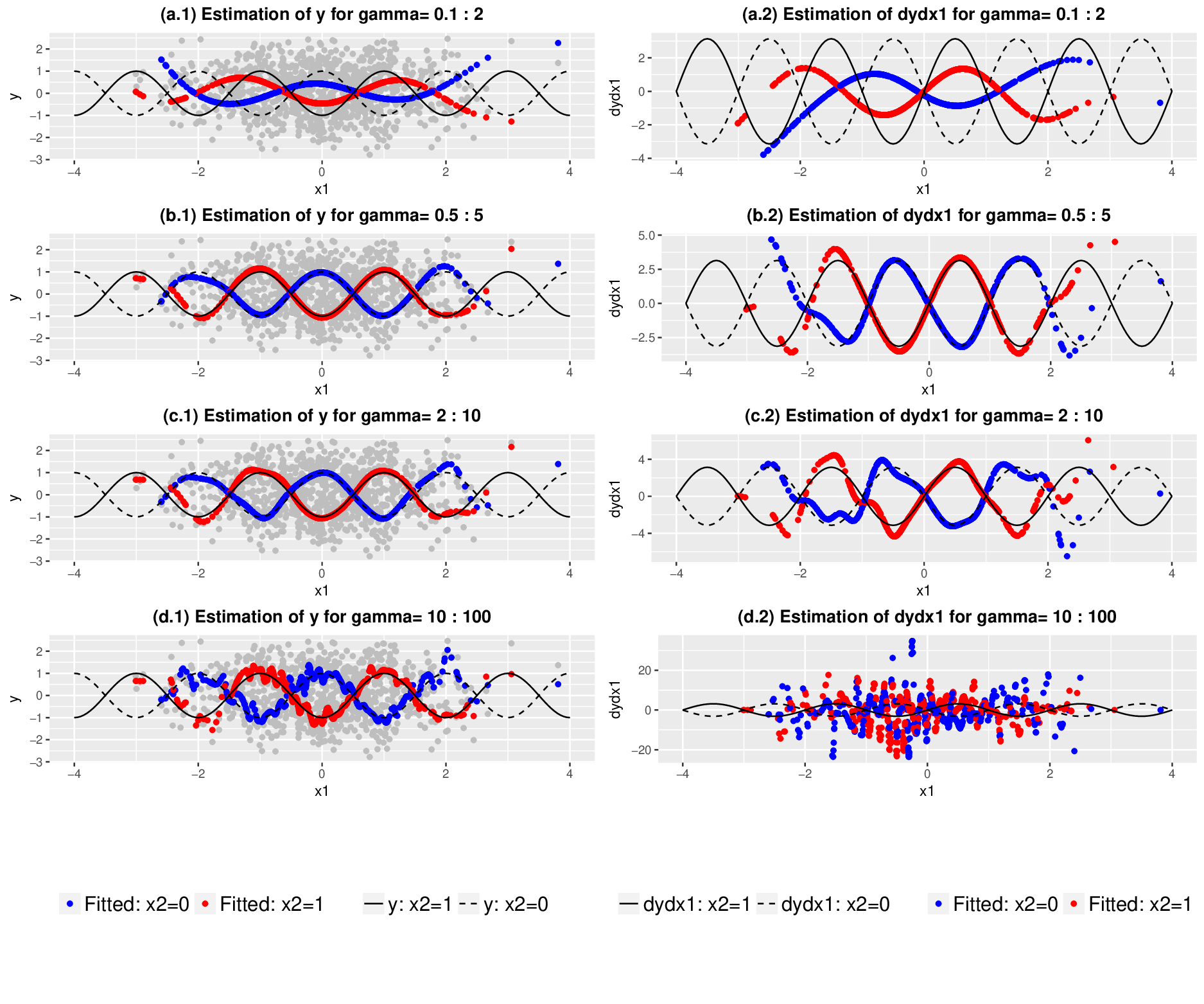}
	\caption{Fitted model for different different ranges of $\gamma$.}
	\label{F:fitsforgs}
\end{figure}

The fitted values and the derivatives are presented in figure \ref{F:fitsforgs}. The first important result is the very poor estimate of the derivatives for large $\gamma$s (between 10 and 100). As we increase the value of $\gamma$, each new split becomes more similar to a CART split, which has no derivative. The fitted values are also noisier for large $\gamma$s. The intervals $[0.5,5]$ and $[2,10]$ are very similar in terms of convergence, but we can see a difference in the estimates of the derivative. The $[2:10]$ interval produced noisier estimates, similar to the cases in which we increased the number of splits in the trees.

The main conclusion from the last three subsections is that the derivatives are much more sensitive to parametrization than fitted values. Good convergence and in-sample fit does not always translate into good estimates for the derivatives. Additionally, over-fitting is more prejudicial to the derivatives than to the fitted values. These small examples indicate that it is better to use small values of $v$, relatively small values of $\gamma$ and grow trees with approximately four splits or less. However, all these parameters together may require a large number of iterations for the BooST algorithm to converge.

\subsection{Monte Carlo Simulation}

In this subsection we show a larger simulation using two DGPs adapted to a smooth version from the XOR and AND DGPs from \citet{wager2014asymptotic}, which we called SXOR and SAND. The DGPs are presented in equations (\ref{eq:XOR}) and (\ref{eq:AND}). The features $x_i$ were sampled from a uniform distribution $U[0,1]$. The first part of the SXOR DGP returns 1 only if $x_1=1$  and $x_2=0$ or $x_1=0$ and $x_2=1$, it will return 0 if $x_1 = x_2 = 0$ or if $x_1=x_2=1$; the second part of the function does the same for $x_3$ and $x_4$. The function is smooth between these points. The AND DGP is a product of logistic functions centered on 0 and have values closer to 1 when all features $x_i$ are also close to 1 and values closer to 0 when all $x_i$ are close to 0.

\begin{equation}
\label{eq:XOR}
    SXOR: y = (0.5x_1^2 + 0.5 x_2^2 + 0.5x_1 + 0.5x_2  - 2x_1x_2) + (0.5x_3^2 + 0.5 x_4^2 + 0.5x_3 + 0.5x_4  - 2x_3x_4) + \varepsilon
\end{equation}

\begin{equation}
\label{eq:AND}
    SAND: y = \prod_{i=1}^4 \frac{1}{1+e^{-x}} + \varepsilon
\end{equation}

We performed 100 simulations from each of these DGPs with sample size $N$ of 300 and 1000 for the BooST, Random Forest, Boosting and a Neural Network. Table \ref{T:Simulation} shows the results for the out-of-sample forecasting errors and the derivative errors in-sample and out-of-sample. The left part of table \ref{T:Simulation} shows that the BooST produces smaller errors for both DGPs. However, the Neural Network also have good results if we consider the case where $N=1000$. Moreover, none of the models perform bad compared to the others in terms of RMSE. The remaining two parts of the table show the results for the derivatives. In this case the BooST performs significantly better than all other models in most cases but the Neural Network also has good results, mostly on large samples, given its smooth nature from the logistic transition. However, the relative difference between the BooST and the Neural Network tends to be bigger for the derivatives than for the conditional mean.

\begin{table}[htb]
\caption{Simulation Study}\label{T:Simulation}
\begin{threeparttable}
\begin{minipage}{\linewidth}
\begin{footnotesize}
The table shows the results for 100 simulations the BooST, the Random Forest, the Boosting and a Neural Network with three hidden layers with 4, 3 and 2 neurons and logistic transition. We estimated the BooST and the Boosting with 200 trees and a step $v$ of 0.05. The $\gamma$ range for the BooST was set to 0.1 to 2 for the SXOR and 0.1 to 1 for the SAND. The Random Forest was set to 500 trees with a minimum of 5 observations per terminal node. The Boosting max depth of each tree and the number of splits per tree in the BooST were both set to 4. The table reports RMSE and BIAS for forecasting $\hat{y}$ and derivatives in-sample and out-of-sample. Derivatives for Random Forests and Boosting were calculated using finite differences setting $h$ to 0.1.
\end{footnotesize}
\end{minipage}
\resizebox{\textwidth}{!}{%
\begin{tabular}{lcccccccccccccccc}
\hline
                     &                           &                  & \multicolumn{4}{c}{Forecasting error $\hat{y}$ Out-of-sample} &  & \multicolumn{4}{c}{$\frac{\partial \hat{y}}{\partial x_1}$ In-sample} &  & \multicolumn{4}{c}{$\frac{\partial \hat{y}}{\partial x_1}$ Out-of-sample} \\ \cline{4-7} \cline{9-12} \cline{14-17} 
                     &                           &                  & BooST     & RF        & Boosting     & NN       &  & BooST           & RF             & Boosting          & NN             &  & BooST            & RF              & Boosting           & NN              \\ \hline
\multirow{4}{*}{SXOR} & \multirow{2}{*}{N = 300}  & RMSE             & 0.149     & 0.218     & 0.190        & 0.184    &  & 0.152           & 0.507          & 0.464             & 0.868          &  & 0.154            & 0.505           & 0.468              & 0.710           \\
                     &                           & BIAS$\times 100$ & 0.002     & 0.135     & 0.661        & 0.064    &  & 0.413           & 0.303          & 0.475             & 3.379          &  & 0.175            & 0.046           & 0.345              & 0.416           \\ \cline{3-7} \cline{9-12} \cline{14-17} 
                     & \multirow{2}{*}{N = 1000} & RMSE             & 0.144     & 0.187     & 0.167        & 0.168    &  & 0.106           & 0.440          & 0.359             & 0.400          &  & 0.107            & 0.437           & 0.363              & 0.352           \\
                     &                           & BIAS$\times 100$ & 0.140     & 0.061     & 0.485        & 0.164    &  & 0.056           & 0.337          & 0.144             & 0.248          &  & 0.072            & 0.388           & 0.771              & 0.244           \\ \hline
\multirow{4}{*}{SAND} & \multirow{2}{*}{N = 300}  & RMSE             & 0.019     & 0.022     & 0.021        & 0.021    &  & 0.015           & 0.032          & 0.041             & 0.019          &  & 0.016            & 0.033           & 0.041              & 0.019           \\
                     &                           & BIAS$\times 100$ & 0.001     & 0.010     & 0.024        & 0.014    &  & 0.064           & 1.848          & 0.848             & 0.518          &  & 0.041            & 1.909           & 0.901              & 0.537           \\ \cline{3-7} \cline{9-12} \cline{14-17} 
                     & \multirow{2}{*}{N = 1000} & RMSE             & 0.019     & 0.020     & 0.020        & 0.019    &  & 0.009           & 0.029          & 0.031             & 0.012          &  & 0.009            & 0.030           & 0.031              & 0.012           \\
                     &                           & BIAS$\times 100$ & 0.022     & 0.039     & 0.021        & 0.022    &  & 0.039           & 1.463          & 0.565             & 0.083          &  & 0.035            & 1.404           & 0.554              & 0.068           \\ \hline
\end{tabular}
}
\end{threeparttable}
\end{table}
\FloatBarrier

\section{Empirical Applications}

\subsection{Engel curve}

In this example, we used data from \cite{delgado1998testing} available in the Ecdat package in R. These data contain 23,971 observations of Spanish households and indicate the proportion of each household's expenditures on food. We removed some extreme outliers, and the final dataset included 23,932 observations. The controls are the total expenditure, the gender of the head of the household, the size of the town and the size of the household. The objective is to estimate the Engel curve, which relates the total expenditure of a family to the proportion of these expenditures on food. The idea is that as we increase the household's expenditures, the food proportion becomes smaller. We estimated the Engel curve and its derivatives and analyzed the results by gender, age and household size. The BooST model was estimated with 1000 trees, $v=0.05$ and $\gamma \in [0.5,5]$. We used a small $v$ because the boosting converges very quickly in this example and we can benefit for a more conservative model.

\begin{figure}[htb]
	\centering
	\includegraphics[scale=0.8]{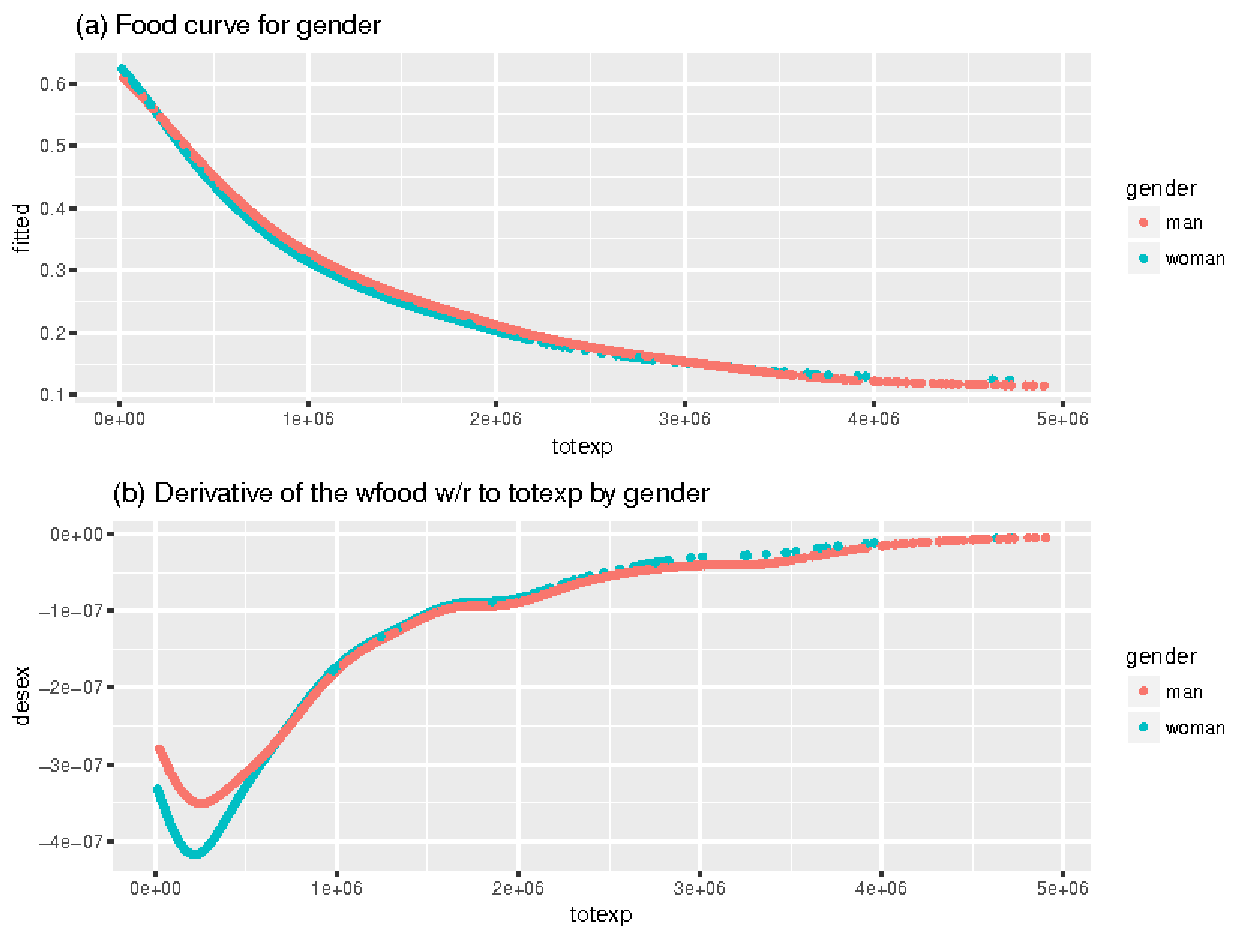}
	\caption{Food curve and its derivative with respect to expenditures by gender}
	\label{F:foodgender}
\end{figure}

Figure \ref{F:foodgender} shows the results for the Engel curve (panel (a)) and its derivative with respect to total expenditures (panel (b)) by gender. The curve has the expected decreasing shape as we increase the total expenditures. Households with less wealth spend approximately 60\% of their income on food. The proportion decreases quickly in the beginning, and the negative slope of the curve becomes increasingly smooth as we reach high levels of expenditures. The derivative reflects the exact same behavior. It is negative in all points and converges to zero as we increase the total expenditures. The difference between gender is not very significant. The curve for women is slightly below the curve for men, but this difference is hardly significant.

\begin{figure}[htb]
	\centering
	\includegraphics[scale=0.8]{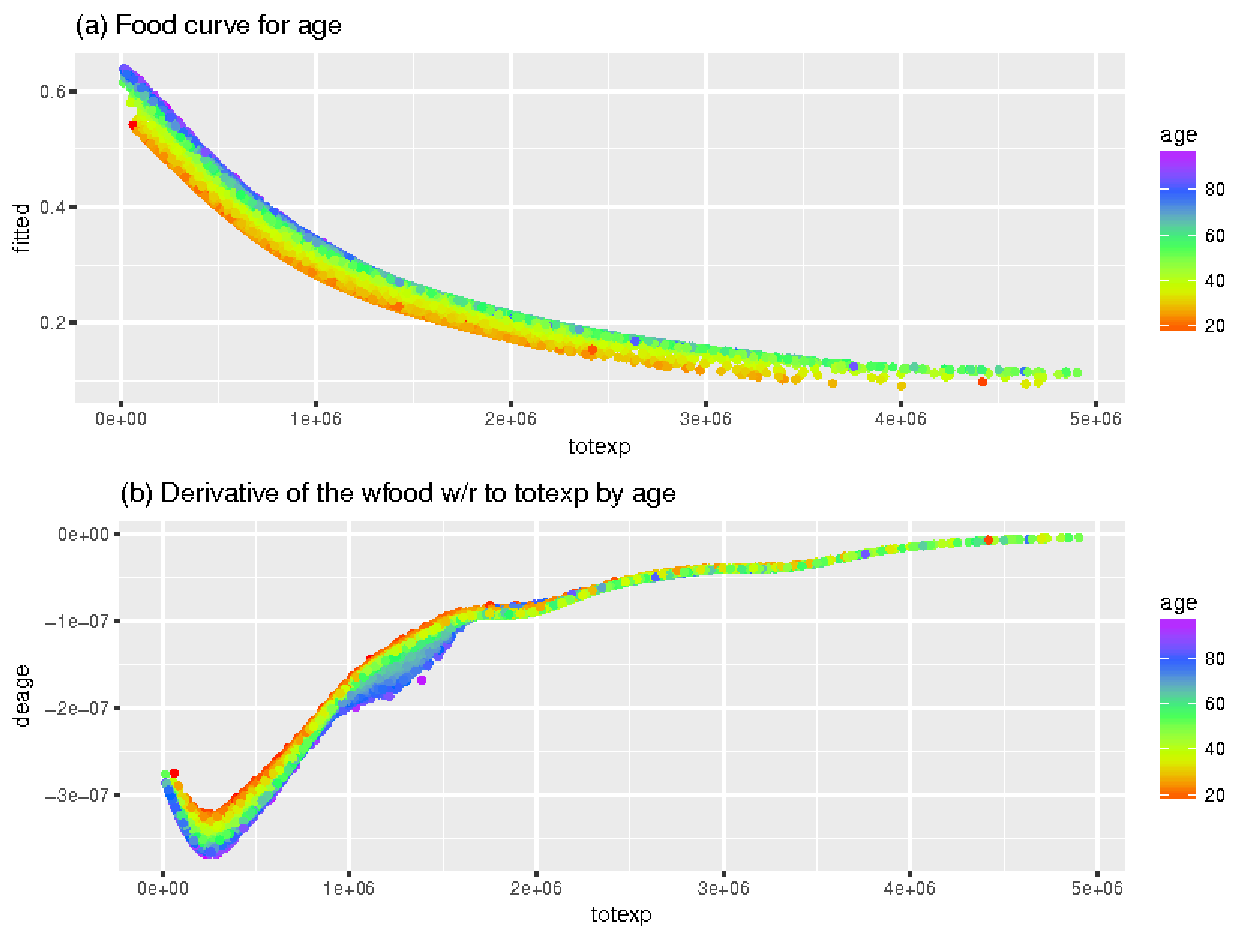}
	\caption{Food curve and its derivative with respect to expenditures by age}
	\label{F:foodage}
\end{figure}

The same results for age of the head of the household are presented in figure \ref{F:foodage}. The figure shows that households managed by younger people usually spend less on food proportionally. Younger people are more likely to be single and have no children. The derivative is more negative for older household heads, but the derivative curve becomes indistinguishable for high levels of expenditures. This result means that households with older managers spend more on food, but this proportion decreases more quickly than in cases in which the household head is young. The indistinguishable derivative for higher expenditures means that once we reach a certain level, the Engel curve has the same slope for all ages.

\begin{figure}[htb]
	\centering
	\includegraphics[scale=0.8]{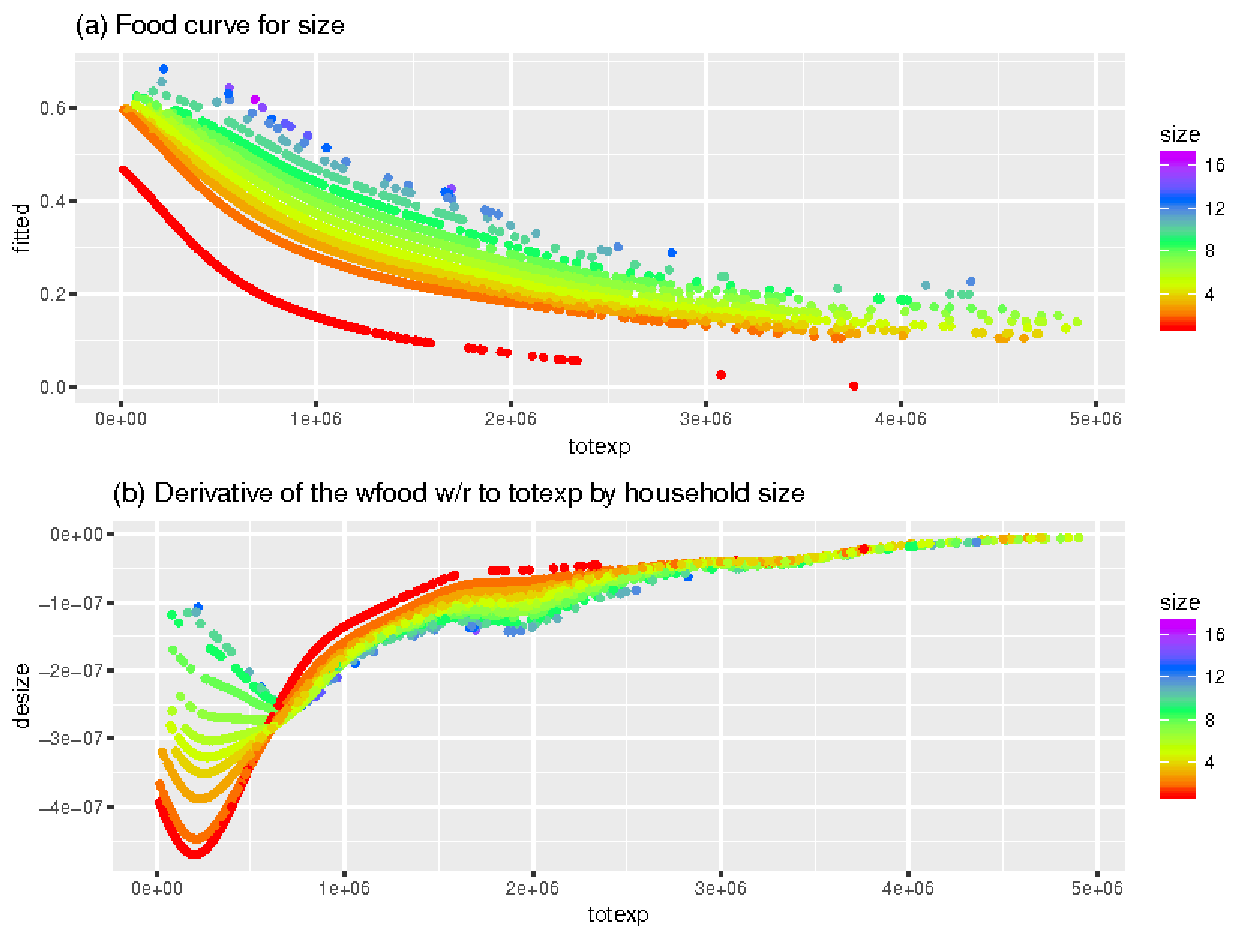}
	\caption{Food curve and its derivative with respect to expenditures by household size}
	\label{F:foodsize}
\end{figure}

The household size in figure \ref{F:foodsize} demonstrates a larger difference between the curves and the derivatives. Small households spend considerably less on food than big households. Moreover, the derivative shows that as the total expenditures increase, the food proportion decreases more quickly for small households and low expenditures. However, this behavior reverses as the total expenditures exceed one million.

\subsection{Housing}

In this empirical example, we will use a house prices dataset scrapped from the web by Tony Pino.\footnote{Available at https://www.kaggle.com/anthonypino/melbourne-housing-market} The data include house transactions in Melbourne, Australia, between 2016 and 2017. After filtering for missing data and extreme outliers, we had 5,926 observations of sold houses. Each house has a selling price and characteristics\footnote{There are other characteristics that we did not use, such as the seller agent, neighborhood (by name), and address.}, such as the number of bedrooms, bathrooms, garages, building area, lot size, distance to the city center, latitude and longitude. Our objective is to estimate the derivatives of prices with respect to building area, latitude and longitude to see how the prices increase as we move to more central areas. The building area derivatives were estimated from a model using all variables mentioned above. We chose to remove the distance variable in the latitude/longitude model because if we change the latitude/longitude, the distance to the center of the city also changes, making it difficult to isolate the effects of each variable. Nevertheless, the results do not change much if we follow this approach: the correlation between the latitude/longitude derivatives obtained from both models is approximately 0.9. The derivative with respect to the building area illustrates the price per extra square meter.

It is realistic to assume that the relation between prices and housing characteristics is highly nonlinear and depends on multiple interactions between the controls. We adopted a conservative strategy and used a shrinkage of $0.05$ with $\gamma$ randomly selected from the $[0.5,5]$ interval. We estimated BooST with 1000 trees for the specifications with and without the distance variable. The derivatives were all estimated for a representative house that has three bedrooms, two bathrooms, two garages, 157 square meters of building area and a lot size of 540 square meters. This setup describes the most common type of house in our dataset. We used this representative house because interpretation becomes much easier if we keep all characteristics constant and focus our analysis on only one variable at a time.

\begin{figure}[htb]
	\centering
	\includegraphics[scale=1]{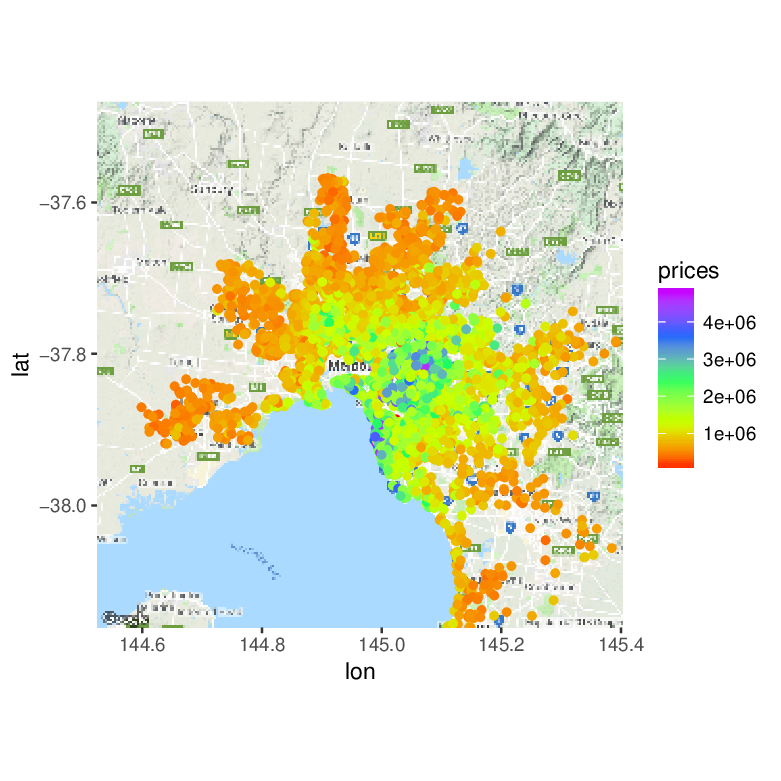}
	\caption{House price data}
	\label{F:houseprices}
\end{figure}

\begin{figure}[htb]
	\centering
	\includegraphics[scale=1]{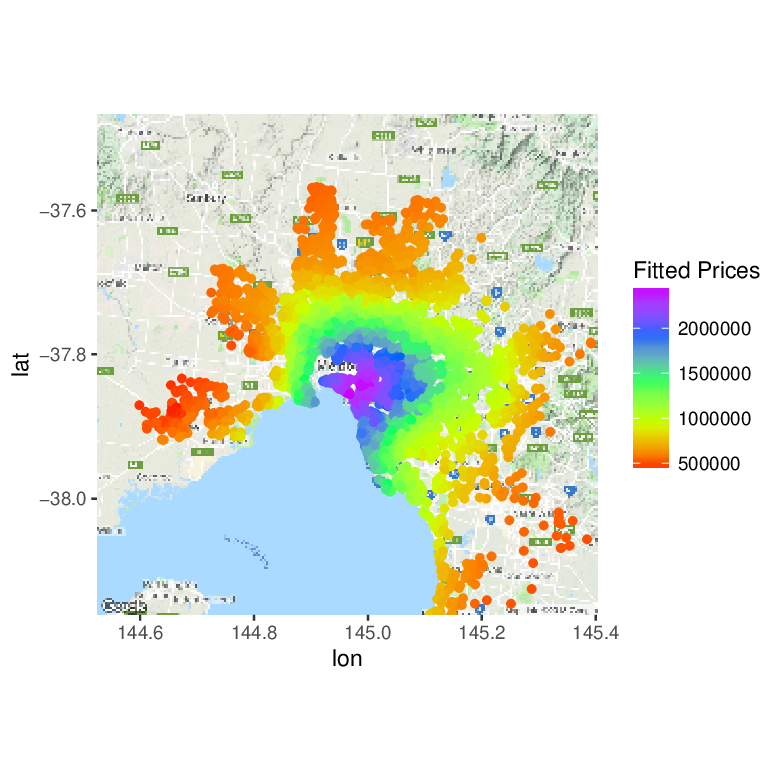}
	\caption{Estimated prices for the representative houses}
	\label{F:housepricesfitted}
\end{figure}

\begin{figure}[htb]
	\centering
	\includegraphics[scale=1]{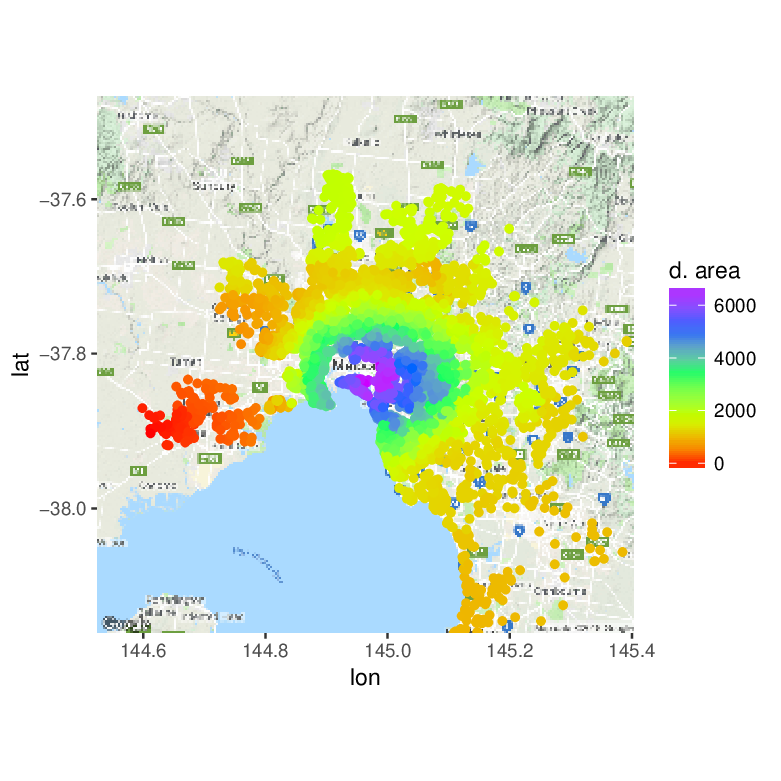}
	\caption{Derivative of house prices with respect to the building area for representative houses}
	\label{F:dearea}
\end{figure}

Figure \ref{F:houseprices} shows the data on a map.\footnote{Available at https://www.kaggle.com/anthonypino/melbourne-housing-market} The color of the dots represents the actual price of the houses sold. The prices increase as we move to the center of the city. The fitted prices for the representative houses are presented in figure \ref{F:housepricesfitted}, which follows the same pattern as the real data but with smaller prices in the right tail of the distribution because we used a representative average-sized house. Patterns are clearer in the fitted model because all representative houses are equal, except by their location. The derivative of prices with respect to the building area is in figure \ref{F:dearea}, which shows the same pattern as the price, i.e., the price of an extra square foot increase as we move to more central areas. The price per square meter goes from close to zero to a little more than 60,00 dollars.

\begin{figure}[htb]
	\centering
	\includegraphics[scale=1]{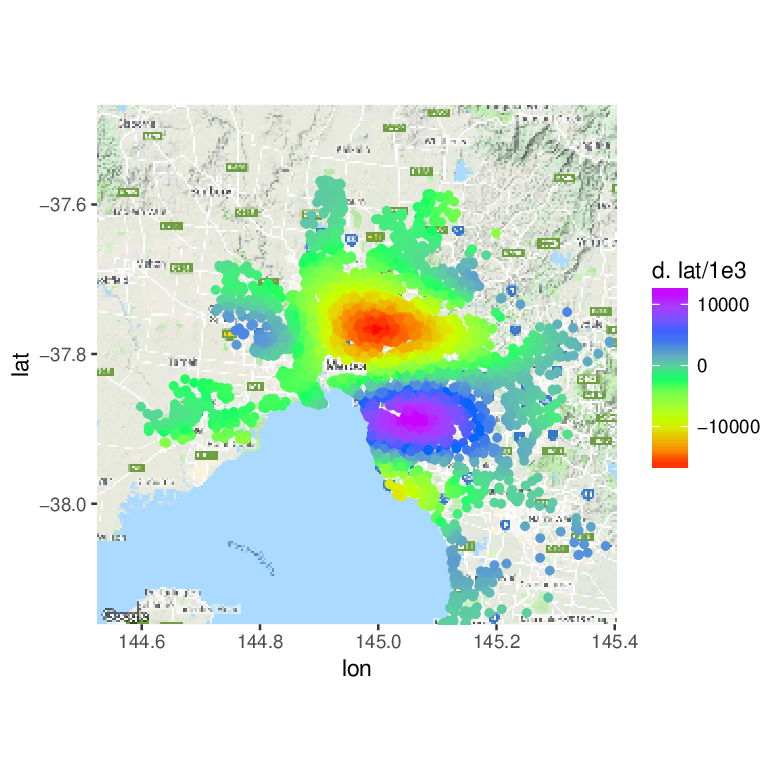}
	\caption{Derivative of house prices with respect to the latitude for representative houses}
	\label{F:delat}
\end{figure}

\begin{figure}[htb]
	\centering
	\includegraphics[scale=1]{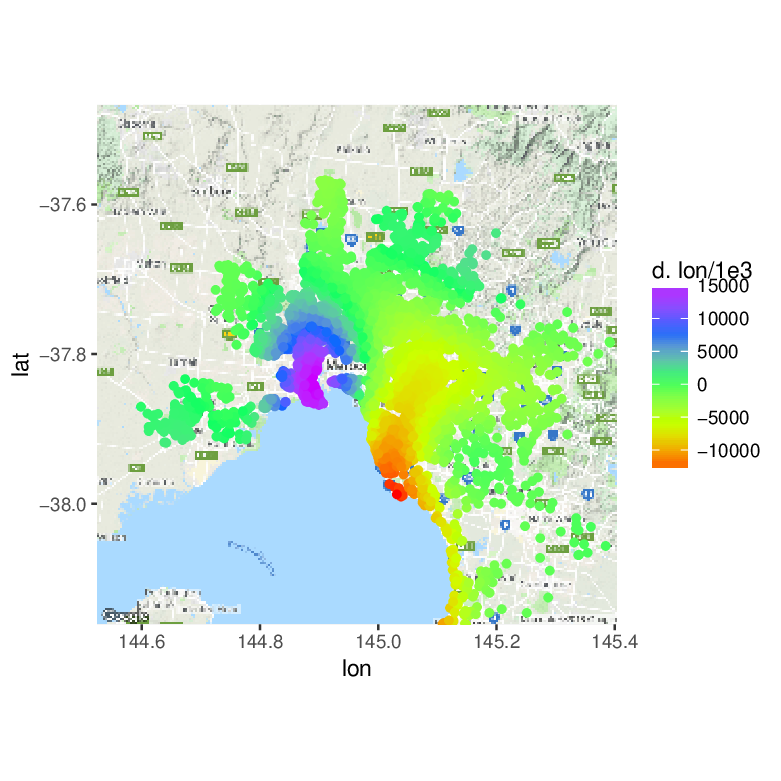}
	\caption{Derivative of house prices with respect to the longitude for representative houses}
	\label{F:delon}
\end{figure}

The latitude and longitude derivatives are presented in figures \ref{F:delat} and \ref{F:delon}, respectively. The results are very interesting. Both figures show many regions in which the derivative is close to zero, which indicates some local optimal points. The latitude figure has a big red area of negative derivatives immediately above the center of the city, indicating the prices reduce a great deal if we move away from the center. The opposite occurs in the south of the central city, showing some large positive values, which indicates that prices increase a great deal if we move towards the center. The longitude figure shows a similar pattern, but from the west to the east. If we are west of the city center, the derivatives are very positive, and they become very negative when we move past the center to the east. Additionally, houses that are not in the central area of the city but are close to the bay also have negative values, showing that prices decrease as we move away from the bay to the east.

This house pricing example can also be used to test how BooST performs in predicting the prices based on the characteristics of the houses. We estimated BooST with three ranges of $\gamma$: $[0.5,5]$, $[2,10]$ and $[5,25]$. All models had four splits in each tree, used a shrinkage of $0.05$, randomly selected $2/3$ of the variables to test in each new split and were estimated with 1000 trees. The experiments were made in a $k$-fold cross-validation scheme with $k=2$, $5$ and $10$. The benchmark models were the unconditional mean, a linear model, a log-linear model using the log of prices, distance, land size and building area with no transformation on the remaining variables. Additionally, we estimated Random Forests\footnote{The Random Forests were estimated with the randomForest package in R.} and boosting\footnote{The boosting was estimated with the xgboost package in R.} with discrete trees for comparison. The boosting was tuned similarly to the BooST with 1000 trees, $v=0.05$, $2/3$ of the variables tested in each new node and a maximum depth of four in each tree, which allows a larger number of splits than the BooST. The Random Forests produced better results when testing only $1/3$ of the variables in each new node and were estimated with 300 trees, more than enough for convergence.

The results are presented in table \ref{T:forecastinghousing}. The boosting was the more accurate model for the 2-fold cross-validation, but the BooST with $\gamma \in [5,25]$ was just slightly less accurate. The same BooST $[5,25]$ was the most accurate model for the 5- and 10-fold cross-validation, but the difference from the discrete boosting was small, approximately 1\% and 2\% relative to the log-linear, respectively. The Random Forest also produced satisfactory results, just slightly worse than those of the BooST $[0.5,5]$. Linear specifications had a poor results with this dataset, as expected.

\begin{table}[htb]
\caption{Cross-validation results for the Melbourne housing dataset}\label{T:forecastinghousing}
\begin{threeparttable}
\begin{minipage}{\linewidth}
\begin{footnotesize}
The table shows the average out-of-sample RMSE of 2-fold, 5-fold and 10-fold cross-validations estimated for all models. The models are the Unconditional Mean, Linear Model, Log-Linear Model, Boosting with CART trees, Random Forest and BooST. The three specifications of BooST are for different ranges of $\gamma$. The folds were randomly generated. All values were divided by the RMSE of the log-linear model, which shows 1 for the three tests. The smallest RMSE in each cross-validation is displayed in bold. Values in parentheses are the p-values for a t-test of each model against the BooST$[5,25]$.
\end{footnotesize}
\end{minipage}
\resizebox{\linewidth}{!}{
\begin{tabular}{lcccccccc}
\hline
           & U-Mean & Linear & LogLinear & Boosting       & R.Forest & BooST$[0.5,5]$ & BooST$[2,10]$ & BooST$[5,25]$ \\ \hline
2-Fold CV  & 1.608   & 1.052   & 1.000     & \textbf{0.735} & 0.780        & 0.766   & 0.754   & 0.736          \\
           & (0.000) & (0.000) & (0.000)   & 0.674          & (0.148)      & (0.065) & (0.315) & -              \\
5-Fold CV  & 1.619   & 1.054   & 1.000     & 0.721          & 0.752        & 0.745   & 0.728   & \textbf{0.711} \\
4          & (0.000) & (0.000) & (0.000)   & 0.918          & (0.101)      & (0.014) & (0.224) & -              \\
10-Fold CV & 1.623   & 1.052   & 1.000     & 0.724          & 0.740        & 0.742   & 0.722   & \textbf{0.706} \\
6          & (0.000) & (0.000) & (0.000)   & 0.512          & (0.244)      & (0.019) & (0.249) & -              \\ \hline
\end{tabular}
}
\end{threeparttable}
\end{table}

\section{Final Remarks}

In this article, we introduce a model that applies the well-known boosting algorithm in Smooth Transition trees (STR-Tree) to estimate derivatives and partial effects in general nonlinear models. The model was named BooST, which stands for Boosted Smooth Transition Regression Trees.

The main contribution of the BooST model is that by using STR-Trees, the estimated model becomes differentiable in all points, and the boosting algorithm makes the estimated derivatives very stable compared to individual trees. The model performed very well in estimating derivatives on simulated data. Additionally, as usual trees, STR-Trees require very little knowledge and assumptions on the data structure. We do not need to make any strong assumptions regarding the type of nonlinearity in the models.

\clearpage


\bibliographystyle{agsm}

\bibliography{sample}

\end{document}